\documentclass{article}

\usepackage{microtype}
\usepackage{graphicx}
\usepackage{subcaption}
\usepackage{booktabs} 
\usepackage{float}

\usepackage{amsmath, amssymb}

\usepackage{hyperref}

\usepackage[accepted]{icml2021}
\usepackage{xcolor}

\icmltitlerunning{Slot Machines: Discovering Winning Combinations of Random Weights in Neural Networks}

\begin{document}
\twocolumn[
\icmltitle{Slot Machines: Discovering Winning Combinations of Random Weights in Neural Networks}




\begin{icmlauthorlist}
\icmlauthor{Maxwell Mbabilla Aladago}{ed}
\icmlauthor{Lorenzo Torresani}{ed}
\end{icmlauthorlist}

\icmlaffiliation{ed}{Department of Computer Science, Dartmouth College, Hanover, NH 03755, USA}

\icmlcorrespondingauthor{Maxwell Mbabilla Aladago}{maxwell.m.aladago.gr@dartmouth.edu}


\vskip 0.3in
]



\printAffiliationsAndNotice{}  

\begin{abstract}
In contrast to traditional weight optimization in a continuous space, we demonstrate the existence of effective random networks whose weights are never updated. By selecting a weight among a fixed set of random values for each individual connection, our method uncovers combinations of random weights that match the performance of traditionally-trained networks of the same capacity. We refer to our networks as ``slot machines'' where each reel (connection) contains a fixed set of symbols (random values). Our backpropagation algorithm ``spins'' the reels to seek ``winning''  combinations, i.e., selections of random weight values that minimize the given loss. Quite surprisingly, we find that allocating just a few random values to each connection (e.g., $8$ values per connection) yields highly competitive combinations despite being dramatically more constrained compared to traditionally learned weights. Moreover, finetuning these combinations often improves performance over the trained baselines.  A randomly initialized VGG-19 with $8$ values per connection contains a combination that achieves $91\%$ test accuracy on CIFAR-10. Our method also achieves an impressive performance of $98.2\%$ on MNIST for neural networks containing only random weights. 
\end{abstract}

\section{Introduction}
\label{introduction}

Innovations in how deep networks are trained have played an important role in the remarkable success deep learning has
produced in a variety of application areas, including image recognition~\citep{resnet-34}, object detection~\citep{ren2015faster, he2017mask}, machine translation~\citep{vaswani2017attention} and language modeling~\citep{brown2020language}. Learning typically involves either optimizing a network from scratch~\citep{Krizhevsky-imagenetclassification}, finetuning a pre-trained model~\citep{yosinski2014transferable} or jointly optimizing the architecture and weights~\citep{zoph2016neural}. Against this predominant background, we pose the following question: can a network instantiated with only random weights achieve competitive results compared to the same model using optimized weights?

For a given task, an untrained, randomly initialized network is unlikely to produce good performance. However, we demonstrate that given sufficient random weight options for each connection, there exist selections of these random weight values that have generalization performance comparable to that of a traditionally-trained network with the same architecture. More importantly, we introduce a method that can find these high-performing randomly weighted configurations {\em consistently} and {\em efficiently}. Furthermore, we show empirically that a small number of random weight options (e.g., $2-8$ values per connection) are sufficient to obtain accuracy comparable to that of the traditionally-trained network. Instead of updating the weights, the algorithm simply selects for each connection a weight value from a fixed set of random weights. 

We use the analogy of ``slot machines'' to describe how our method operates. Each reel in a Slot Machine has a fixed set of symbols. The reels are jointly spinned in an attempt to find winning combinations. In our context, each connection has a fixed set of random weight values. Our algorithm ``spins the reels'' in order to find a winning combination of symbols, i.e., selects a weight value for each connection so as to produce an instantiation of the network that yields strong performance. While in physical Slot Machines the spinning of the reels is governed by a fully random process, in our Slot Machines the selection of the weights is guided by a method that optimizes the given loss at each spinning iteration.

More formally, we allocate $K$ fixed random weight values to each connection. Our algorithm assigns a quality score to each of these $K$ possible values. In the forward pass a weight value is selected for each connection based on the scores. The scores are then updated in the backward pass via stochastic gradient descent. However, the weights are never changed. By evaluating different combinations of fixed randomly generated values, this extremely simple procedure finds weight configurations that yield high accuracy.

We demonstrate the efficacy of our algorithm through experiments on MNIST and CIFAR-10. On MNIST, our randomly weighted Lenet-300-100~\citep{lecun1998lenet} obtains a $97.0\%$ test set accuracy when using $K=2$ options per connection and $98.2\%$ with $K=8$. On CIFAR-10~\citep{Krizhevsky09learningmultiple}, our six layer convolutional network outperforms the traditionally-trained network when selecting from $K=8$ fixed random values at each connection. 
 
 Finetuning the models obtained by our procedure generally boosts performance over networks with optimized weights albeit at an additional compute cost (see Figure~\ref{fig:fine-tuning}). Also, compared to traditional networks, our networks are less memory efficient due to the inclusion of scores. That said, our work casts light on some intriguing phenomena about neural networks:

  \begin{itemize}
     \item First, our results suggest a performance comparability between selection from multiple random weights and traditional training by continuous weight optimization. This underscores the effectiveness of strong initializations. 
     \item Second, this paper further highlights the enormous expressive capacity of neural networks. ~\cite{maennel2020neural} show that contemporary neural networks are so powerful that they can memorize randomly generated labels. This work builds on that revelation and demonstrates that current networks can model challenging non-linear mappings extremely well even by simple selection from random weights.
     \item This work also connects to recent observations~\citep{malach2020proving, frankle2018lottery} suggesting strong performance can be obtained by utilizing gradient descent to uncover effective subnetworks.
     \item  Finally, we are hopeful that our novel model ---consisting in the introduction of multiple weight options for each edge--- will inspire other initialization and optimization strategies. 
 \end{itemize}
 
 \begin{figure*}[!t]
    \centering
    \begin{subfigure}{0.22\linewidth}
    \includegraphics[width=\linewidth]{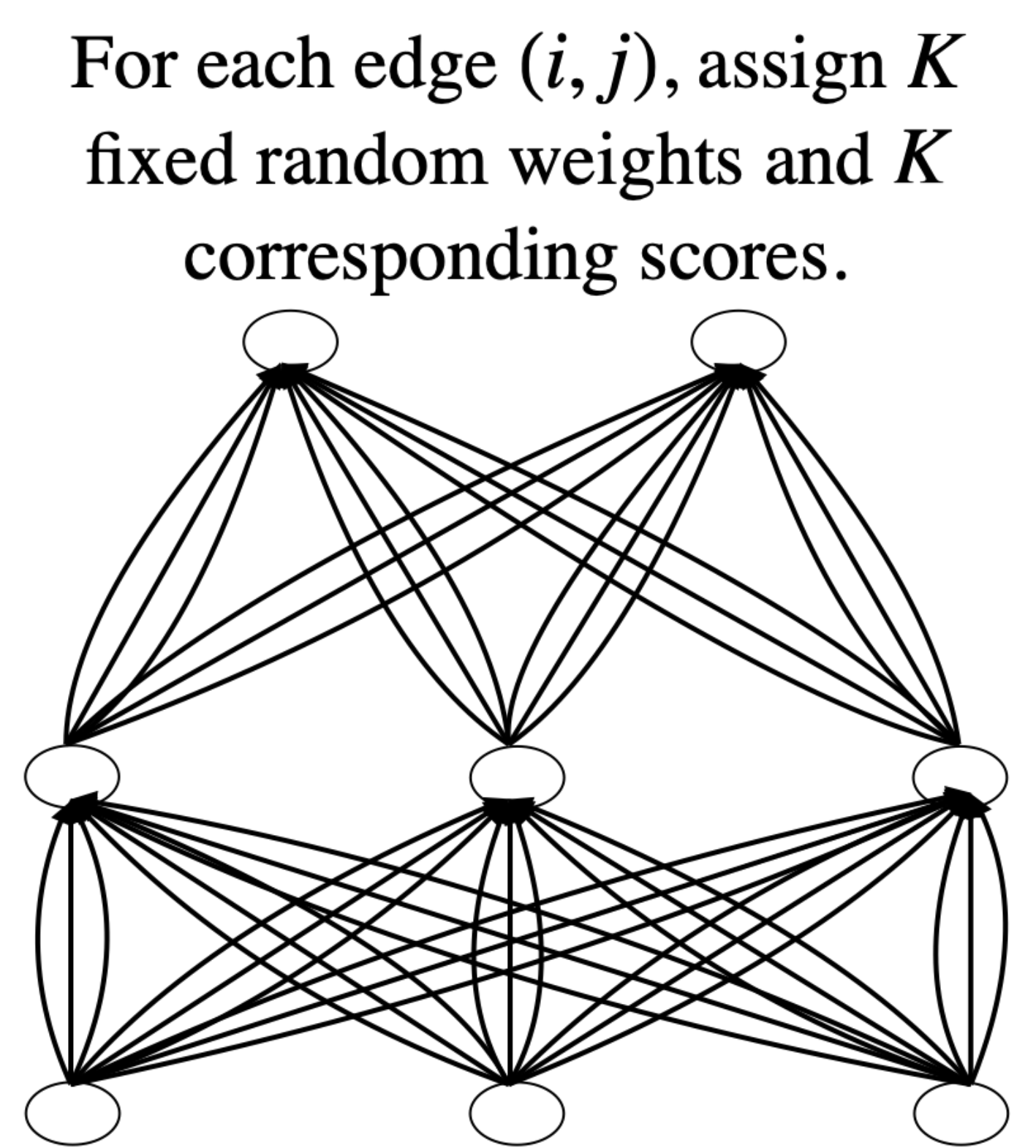}
    \end{subfigure}
    \hspace{.7cm}
    \begin{subfigure}{0.22\linewidth}
    \includegraphics[width=\linewidth]{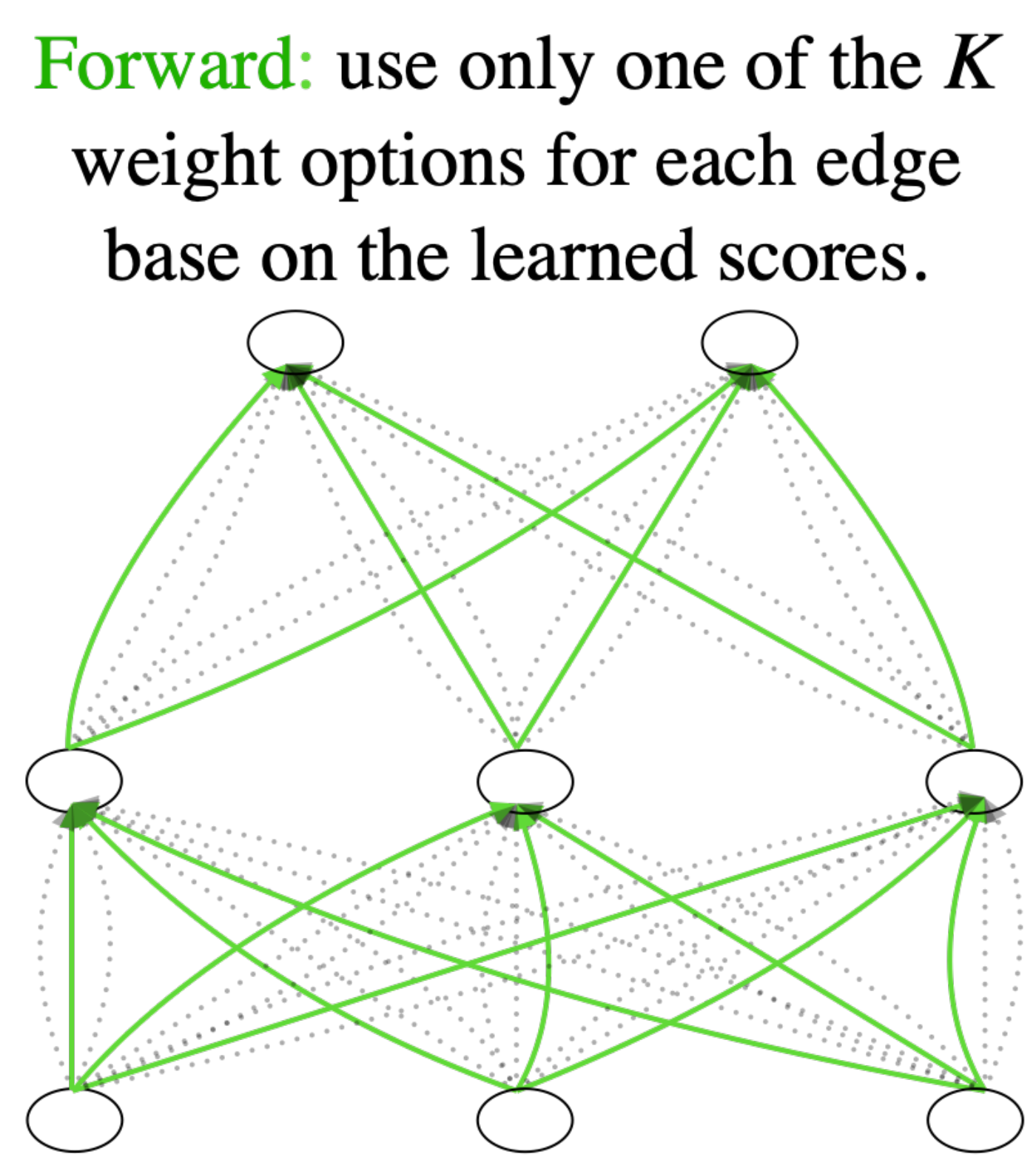}
    \end{subfigure}
    \hspace{.7cm}    
    \begin{subfigure}{0.22\linewidth}
    \includegraphics[width=\linewidth]{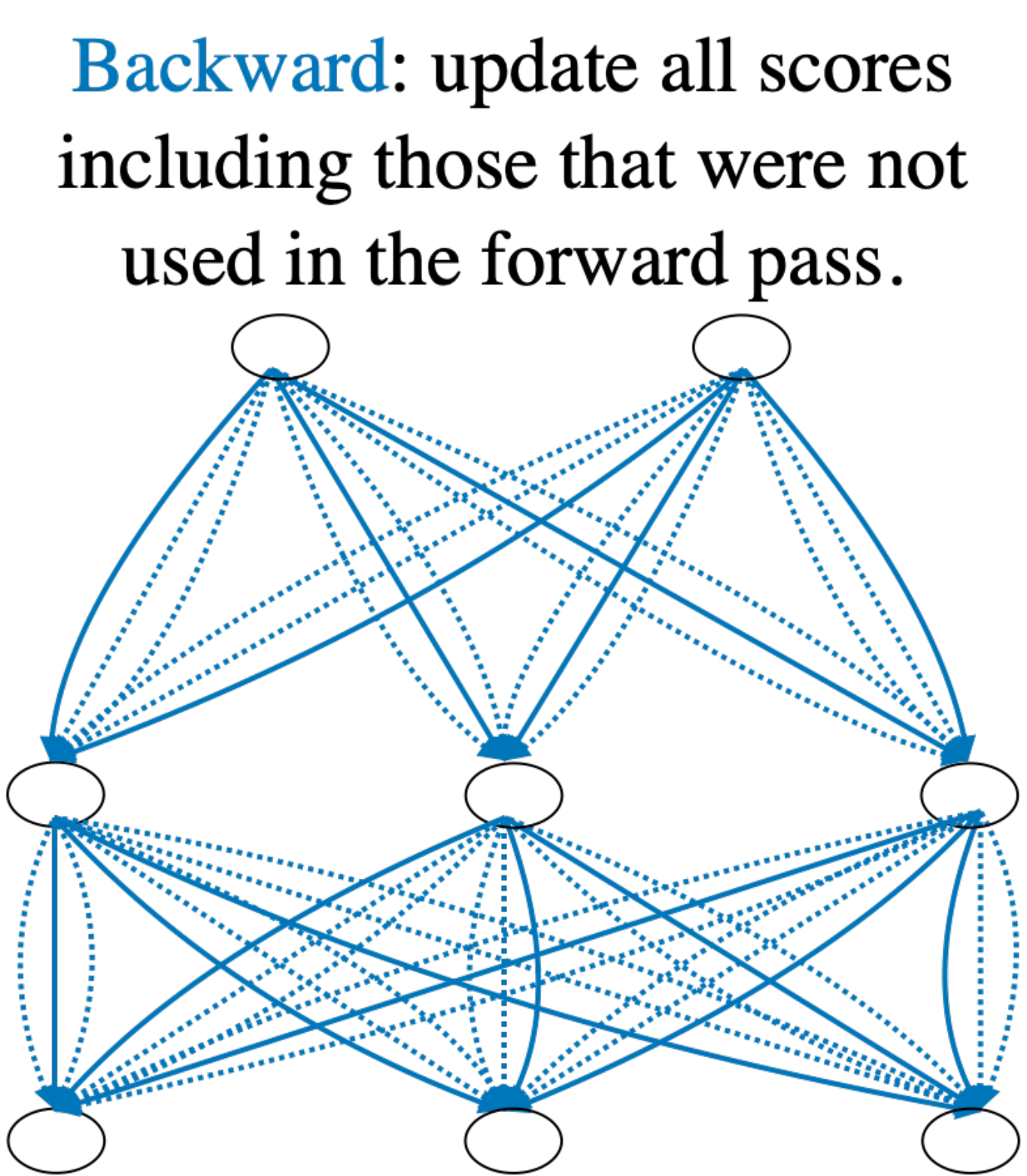}
    \end{subfigure}
    \caption{Our method assigns a set of $K$ random weight options to each connection (we use $K=3$ in this illustration). During the forward pass, one of the $K$ values is selected for each connection, based on a quality score computed for each weight value. On the backward pass, the quality scores of all weights are updated using a straight-through gradient estimator~\citep{bengio2013estimating}, enabling the network to sample better weights in future passes. Unlike the scores, the weights are never changed.}
    \label{fig:algorithm}
\end{figure*}

\section{Related Work}
\label{related-work}
\textbf{Supermasks and the Strong Lottery Ticket Conjecture. } The lottery ticket hypothesis was articulated in~\citep{frankle2018lottery} and states that a randomly initialized neural network contains sparse subnetworks which when trained in isolation from scratch can achieve accuracy similar to that of the trained dense network. Inspired by this result, \citet{zhou2019deconstructing} present a method for identifying subnetworks of randomly initialized neural networks that achieve better than chance performance without training. These subnetworks (named ``supermasks") are found by assigning a probability value to each connection. These probabilities are used to sample the connections to use and are updated via stochastic gradient descent. Without ever modifying the weights, \citet{zhou2019deconstructing} find subnetworks that perform impressively across multiple datasets. 

 Follow up work by \citet{ramanujan2019whats} finds supermasks that match the performance of a dense network. On ImageNet~\citep{imagenet-cvpr09}, they find subnetworks within a randomly weighted ResNet-50~\citep{zagoruyko2016wide} that match the performance of a smaller, trained ResNet-34~\citep{resnet-34}. 
 
 Accordingly, they propose the strong lottery ticket conjecture: a sufficiently overparameterized, randomly weighted neural network contains a subnetwork that performs as well as a traditionally-trained network with the same number of parameters. \citet{ramanujan2019whats} adopts a deterministic protocol in their so-called ``edge-popup" algorithm for finding supermasks instead of the stochastic algorithm  of~\citet{zhou2019deconstructing}.  
 
 These empirical results as well recent theoretical ones~\citep{malach2020proving, pensia2020optimal} suggest that pruning a randomly initialized network is just as good as optimizing the weights, provided a good pruning mechanism is used. Our work corroborates this intriguing phenomenon but differs from these prior methods in a significant aspect. We eliminate pruning completely and instead introduce multiple weight values per connection. Thus, rather than selecting connections to define a subnetwork, our method selects weights for all connections in a network of fixed structure. 
 Although our work has interesting parallels with pruning, it is different from pruning as all connections remain active in every forward pass. 
 
 \textbf{Pruning at Initialization. } The lottery ticket hypothesis also inspired several recent work aimed towards pruning  (i.e., predicting ``winning" tickets) at initialization~\citep{lee2019signal, lee2018snip, tanaka2020pruning,  wang2020picking}. 
 
Our work is different in motivation from these methods and those that train only a subset of the weights~\citep{hoffer2018fix, amir2019intri}. Our aim is to find neural networks with random weights that match the performance of traditionally-trained networks with the same number of parameters. 

\textbf{Weight Agnostic Neural Networks. } ~\citet{gaier2019weight} build neural network architectures with high performance in a setting where all the weights have the same shared random value. The optimization is instead performed over the architecture~\citep{stanley2002neat}.

 They show empirically that the network performance is indifferent to the shared value but defaults to random chance when all the weights assume different random values. Although we do not perform weight training, the weights in this work have different random values. Further, we build our models using fixed  architectures.

\textbf{Low-bit Networks and Quantization Methods. } As in binary networks~\citep{CourbariauxB16, RastegariORF16} and network quantization~\citep{hubara2017quantized, Wang2018}, the parameters in slot machines are drawn from a finite set. However, whereas the primary objective in quantized networks is mostly compression and computational speedup, the motivation behind slot machines is recovering good performance from randomly initialized networks. Accordingly, slot machines use real-valued weights as opposed to the binary (or integer) weights used by low-bit networks. Furthermore, the weights in low-bit networks are usually optimized directly whereas only associated scores are optimized in slot machines.

\textbf{Random Decision Trees. } Our approach is inspired by the popular use of random subsets of features in the construction of decision trees~\citep{Breiman84}. Instead of considering all possible choices of features and all possible splitting tests at each node, random decision trees are built by restricting the selection to small random subsets of feature values and splitting hypotheses. We adapt this strategy to the training of neural network by restricting the optimization of each connection over a random subset of weight values.

\section{Slot Machines: Networks with Fixed Random Weight Options}
\label{method}

Our goal is to construct non-sparse neural networks that achieve high accuracy by selecting a value from a fixed set of completely random weights for each connection. We start by providing an intuition for our method in Section~\ref{intuition}, before formally defining our algorithm in Section~\ref{main-algorithm} .

\subsection{Intuition}\label{intuition}

An untrained, randomly initialized network is unlikely to perform better than random chance. Interestingly, the impressive advances of \citet{ramanujan2019whats} and \citet{zhou2019deconstructing} demonstrate that networks with random weights can in fact do well if pruned properly. In this work, instead of pruning we explore weight selection from fixed random values as a way to obtain effective networks. To provide an intuition for our method, consider an untrained network $N$ with one weight value for each connection, as typically done. If the weights of $N$ are drawn randomly from an appropriate distribution $\mathcal{D}$ (e.g., ~\citet{pmlr-v9-glorot10a} or ~\citet{He2050}), there is an extremely small but nonzero probability that ${N}$ obtains good accuracy (say, greater than a threshold $\tau$) on the given task. Let $q$ denote this probability. Also consider another untrained network $N_K$ that has the same architectural configuration as $N$ but with $K>1$ weight choices per connection. If $n$ is the number of connections in $N$, then $N_K$ contains within it $K^n$ different network instantiations that are architecturally identical to $N$ but that differ in weight configuration. If the weights of $N_K$ are sampled from $\mathcal{D}$, then the probability that none of the 
$K^n$ networks obtains good accuracy is essentially $(1 - q)^{K^n}$. This probability decays quickly as either $K$ or $n$ increases. 
Our method finds randomly weighted networks that achieve very high accuracy even with small values of $K$. For instance, a six layer convolutional network with $2$ random values per connection obtains $85.1\%$ test accuracy on CIFAR-10.

But how do we select a good network from these $K^n$ different networks? Brute-force evaluation of all possible configurations is clearly not feasible due to the massive number of different hypotheses. Instead, we present an algorithm, shown in Figure~\ref{fig:algorithm}, that iteratively searches the best combination of connection values for the entire network by optimizing the given loss. To do this, the method learns a real-valued quality score for each weight option. These scores are used to select the weight value of each connection during the forward pass. The scores are then updated in the backward pass based on the loss value in order to improve training performance over iterations.

\subsection{Learning in Slot Machines}\label{main-algorithm}
Here we introduce our algorithm for the case of fully-connected networks but the description extends seamlessly to convolutional networks. A fully-connected neural network is an acyclic graph consisting of a stack of $L$ layers $[1, \cdots, L]$ where the $\ell$th layer has $n_\ell$ neurons.
The activation $h(x)^{(\ell)}_i$ of neuron $i$ in layer $\ell$ is given by

\begin{equation}
    h(x)^{(\ell)}_i = g\left(\sum_{j = 1}^{n_{\ell - 1}}h(x)_j^{(\ell - 1)}W^{(\ell)}_{ij}\right)
\end{equation}

where $W^{(\ell)}_{ij}$ is the weight of the connection between neuron $i$ in layer $\ell$ and neuron $j$ in layer $\ell - 1$,  $x$ represents the input to the network, and $g$ is a non-linear activation function. Traditionally, $W^{(\ell)}_{ij}$ starts off as a random value drawn from an appropriate distribution before being optimized with respect to a dataset and a loss function using gradient descent. In contrast, our method does not ever update the weights. Instead, it associates a set of $K$ possible weight options for each connection\footnote{For simplicity, we use the same number of weight options $K$ for all connections in a network.}, and then it optimizes the selection of weights to use from these predefined sets for all connections. 
 
 \noindent \textbf{Forward Pass. } Let $\{W_{ij1}, \hdots, W_{ijK}\}$\footnote{For brevity, from now on we omit the superscript denoting the layer.} be the set of the $K$ possible weight values for connection ($i, j$) and let $s_{ijk}$ be the ``quality score" of value $W_{i,j,k}$, denoting the preference for this value over the other possible $K-1$ values.  We define a selection function $\rho$ which takes as input the scores $\{s_{ij1}, \hdots, s_{ijK}\}$ and returns an index between $1$ and $K$. In the forward pass, we set the weight of ($i, j$) to $W_{ijk^*}$ where $k^* = \rho(s_{ij1}, \hdots, s_{ijK})$.

In our work, we set $\rho$ to be either the $\arg \max$ function (returning the index corresponding to the largest score) or the sampling from a Multinomial distribution defined by $\{s_{ij1}, \hdots, s_{ijK}\}$.  We refer to the former as Greedy Selection ($\textsc{GS}$). We name the latter Probabilistic Sampling ($\textsc{PS}$) and implement it as
    \begin{equation}
        \rho  \sim \text{Mult}\left(\frac{e^{s_{ij1}}}{\sum_{k = 1}^K e^{s_{ijk}}}, \cdots, \frac{e^{s_{ijK}}}{\sum_{k = 1}^K e^{s_{ijk}}}\right)
    \end{equation} 
    where $\text{Mult}$ is the multinomial distribution. 
    The empirical comparison between these two selection strategies is given in Section ~\ref{selection-method}. 
    
    We note that, although $K$ values per connection are considered during training (as opposed to the infinite number of possible values in traditional training), only one value per connection is used at test time. The final network is obtained by selecting for each connection the value corresponding to the highest score (for both \textsc{GS} and \textsc{PS}) upon completion of training. Thus, the effective capacity of the network at inference time is the same as that of a traditionally-trained network.

\textbf{Backward Pass. } In the backward pass, all the scores are updated with straight-through gradient estimation since $\rho$ has a zero gradient almost everywhere. The straight-through gradient estimator~\citep{bengio2013estimating} treats $\rho$ essentially as the identity function in the backward pass by setting the gradient of the loss with respect to $s_{ijk}$ as
 \begin{equation}
     \nabla_{s_{ijk}} \gets \frac{\partial \mathcal{L}}{\partial a(x)^{(\ell)}_i}h(x)^{(\ell - 1)}_jW^{(\ell)}_{ijk} \ 
 \end{equation}
 for $k \in \{1,\cdots, K\}$ where $\mathcal{L}$ is the objective function. $a(x)_i^{(\ell)}$ is the pre-activation of neuron $i$ in layer $\ell$. 
 Given $\alpha$ as the learning rate, and ignoring momentum, we update the scores via stochastic gradient descent as  
 \begin{equation}
     \tilde s_{ijk} = s_{ijk} - \alpha \nabla_{s_{ijk}}
 \end{equation}
 where $\tilde s_{ijk}$ is the score after the update. Our experiments demonstrate that this simple algorithm learns to select effective configurations of random weights resulting in impressive results across different datasets and models. 
 

 \begin{table*}[!t]
    \centering
    \caption{Architecture specifications of the networks used in our experiments. The Lenet network is trained on MNIST. The CONV-$x$ models are the same VGG-like architectures used in~\citep{frankle2018lottery, zhou2019deconstructing, ramanujan2019whats}. All convolutions use $3 \times 3$ filters and pool denotes max pooling.}
    \label{tab:architecture}
    \vskip 0.15in
    {\scriptsize \begin{tabular}{lccccc}
         \toprule
         \emph{Network} & Lenet & CONV-2 & CONV-4 & CONV-6 & VGG-19\\
          \midrule 
           & & & & &   2x64, pool\\
            & & & & &   2x128, pool\\
           & & & &  $64, 64$, pool &  2x256, pool\\
          & & &  $64, 64$, pool &  $128, 128$, pool & 4x512, pool\\
         \emph{Convolutional Layers} & & $64, 64$, pool & $128, 128$, pool & $256, 256$, pool & 4x512, avg-pool\\
         \midrule
         \emph{Fully-connected Layers} & $300,100, 10$ & $256, 256, 10$ & $256, 256, 10$ & $256, 256, 10$ & 10\\
         \midrule
         \emph{Epochs: Slot Machines} & $200$ &  $200$ &  $200$ &  $200$ & $220$\\
         \emph{Epochs: Learned Weights} & $200$ &  $200$ &  $330$ &  $330$ & $320$\\
         \bottomrule
    \end{tabular}\vspace{.5cm}}
\end{table*}

\begin{figure*}[!t]
    \centering
    (a)
    \begin{subfigure}{0.22\linewidth}
    \includegraphics[width=\linewidth]{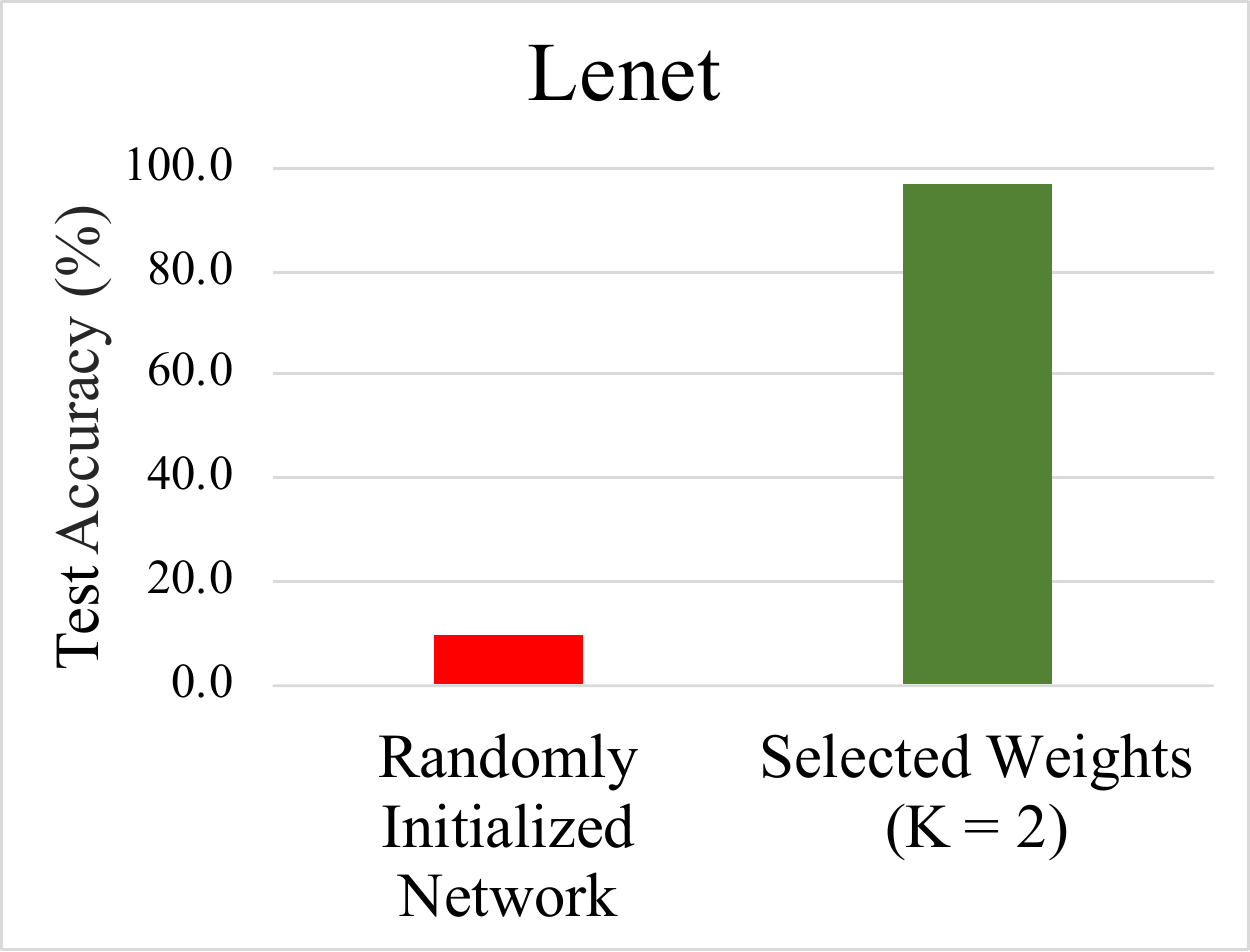}
    \end{subfigure}
    ~~~~(b)
    \begin{subfigure}{0.22\linewidth}
    \includegraphics[width=\linewidth]{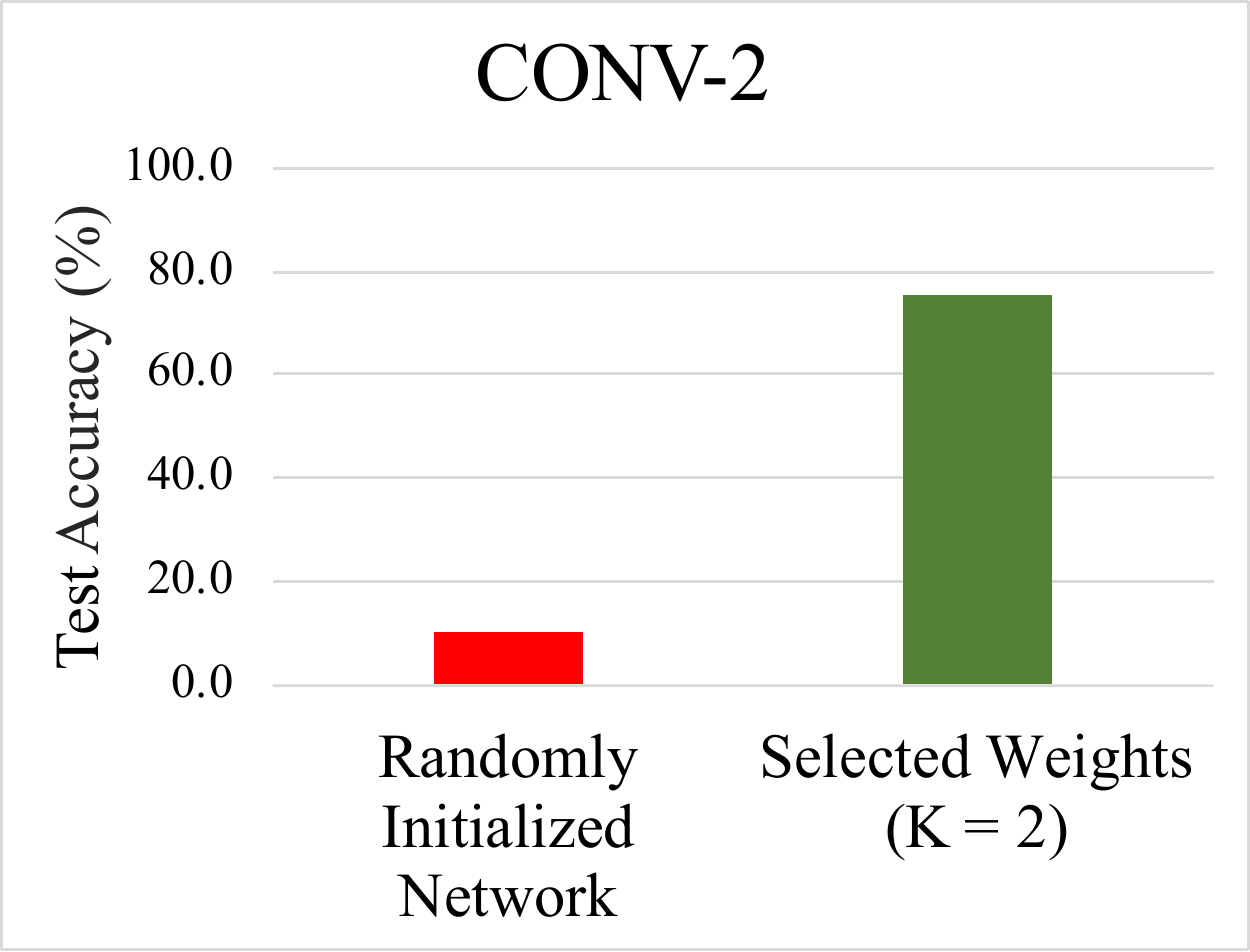}
    \end{subfigure}
    \begin{subfigure}{0.22\linewidth}
    \includegraphics[width=\linewidth]{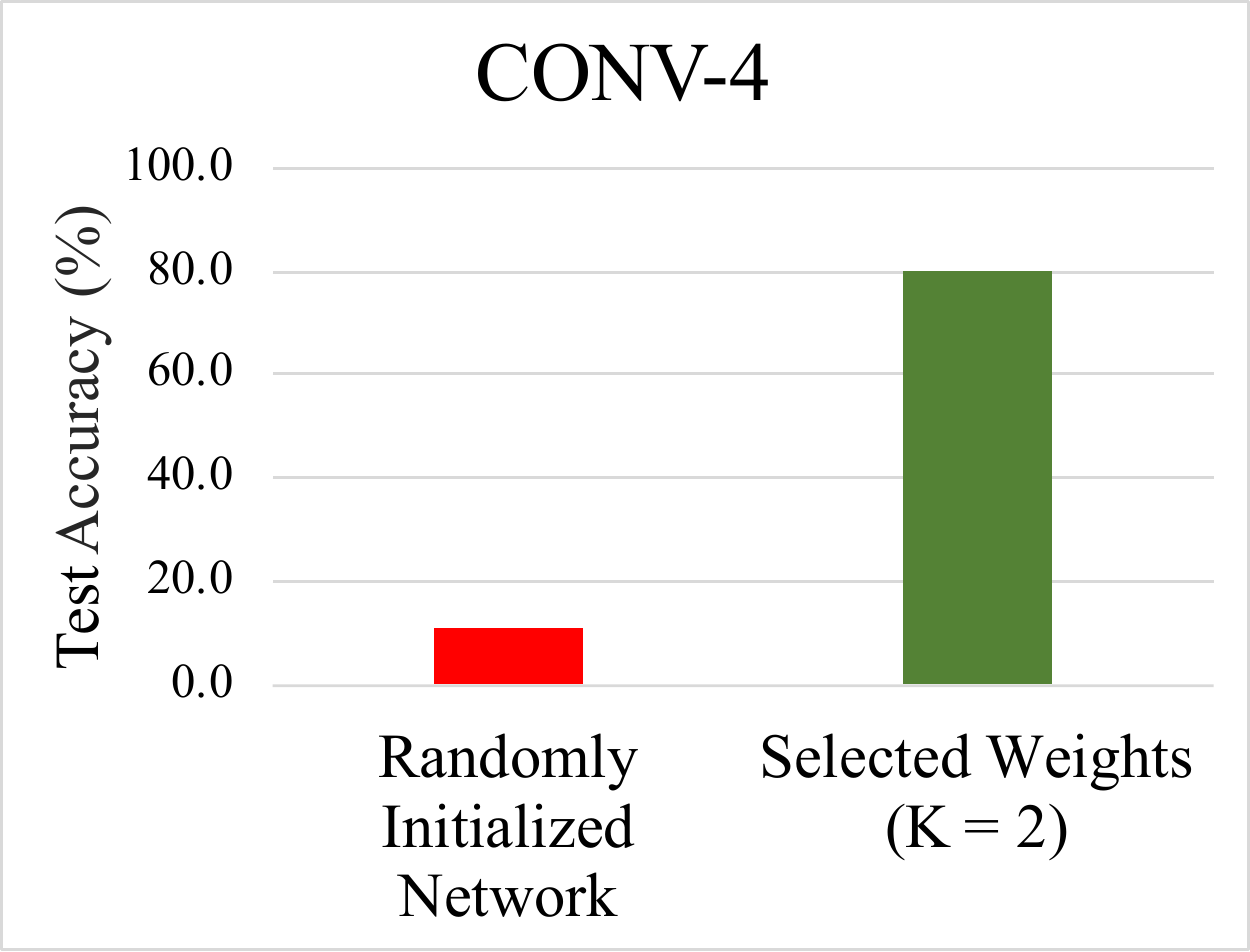}
    \end{subfigure}
    \begin{subfigure}{0.22\linewidth}
    \includegraphics[width=\linewidth]{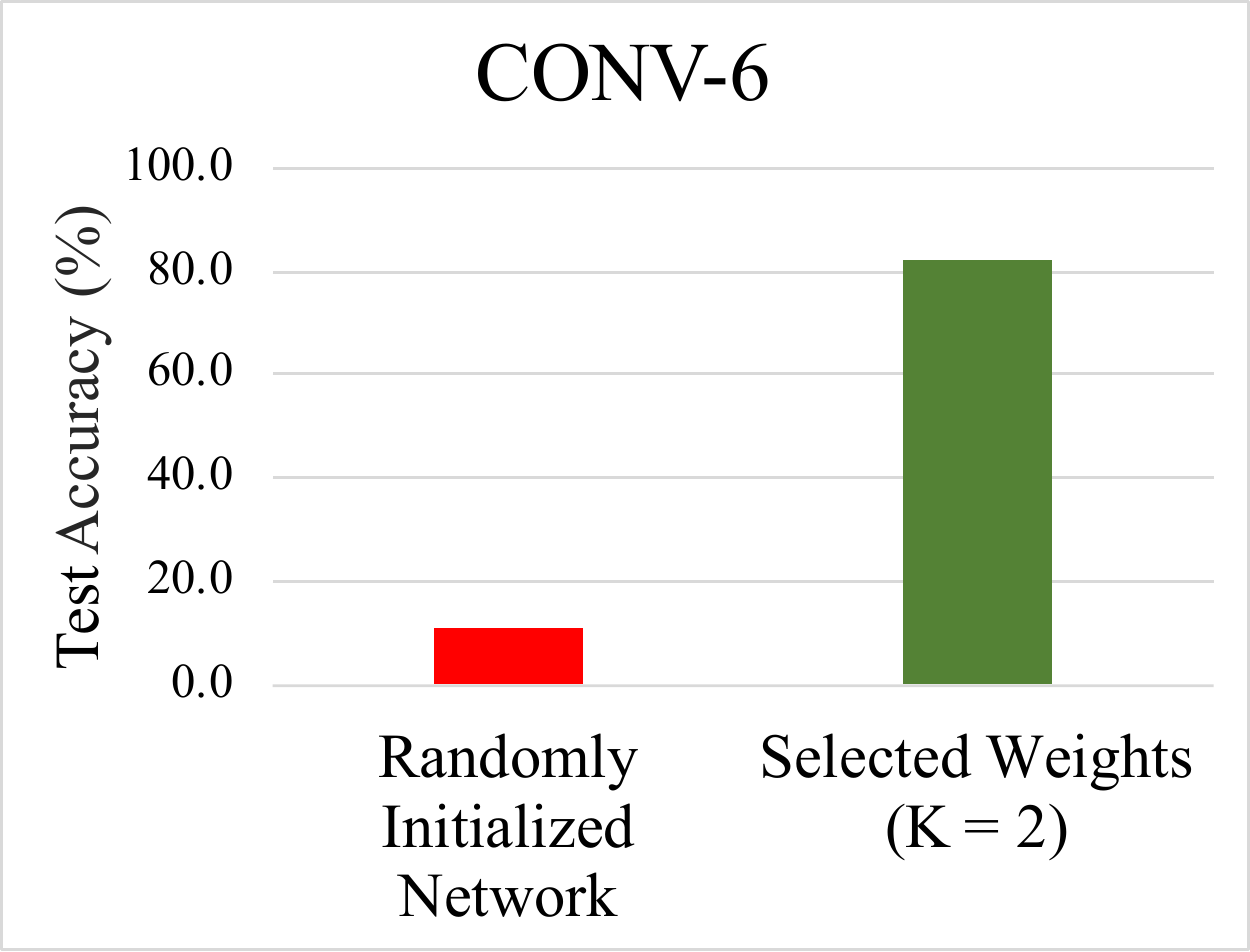}
    \end{subfigure}
    \caption{Selecting from only $K=2$ weight options per connection already dramatically improves accuracy compared to an untrained network that performs at random chance ($10\%$) on both (a) MNIST and (b) CIFAR-10. The first bar in each plot shows the performance of an untrained randomly initialized network and the second bar shows the results of selecting random weights with $\textsc{GS}$ using $K=2$ options per connection.}
    \label{fig:comparision-random}
\end{figure*}


 \section{Experiments}
\label{experiments}
\subsection{Experimental Setup}\label{setup}
\noindent The weights of all our networks are sampled uniformly at random from a Glorot Uniform distribution~\citep{pmlr-v9-glorot10a}, $\mathbb{U}(-\sigma_x, \sigma_x)$ where $\sigma_x$ is the standard deviation of the Glorot Normal distribution. We ignore $K$, the number of options per connection, when computing the standard deviation since it does not affect the network capacity in the forward pass. Like for the weights, we initialize the scores independently from a uniform distribution $\mathbb{U}(0, \lambda\sigma_x)$ where  $\lambda$ is a small constant. We use $\lambda = 0.1$ for all fully-connected layers and set $\lambda$ to $1$ when initializing convolutional layers. We use $15\%$ and $10\%$ of the training sets of MNIST  and CIFAR-10, respectively, for validation. We report performance on the separate test set. On MNIST, we experiment with the Lenet-300-100~\citep{lecun1998lenet} architecture following the protocol in~\citet{frankle2018lottery}. We also use the VGG-like architectures used thereof and in \citet{zhou2019deconstructing} and \citet{ramanujan2019whats}. We denote these networks as CONV-2, CONV-4, and CONV-6. These architectures are provided in Table~\ref{tab:architecture} for completeness. All our plots show the averages of four different independent trials. Error bars whenever shown are the minimum and maximum over the trials. 


 \begin{figure*}[!h]
    \centering
    \includegraphics[width=\linewidth]{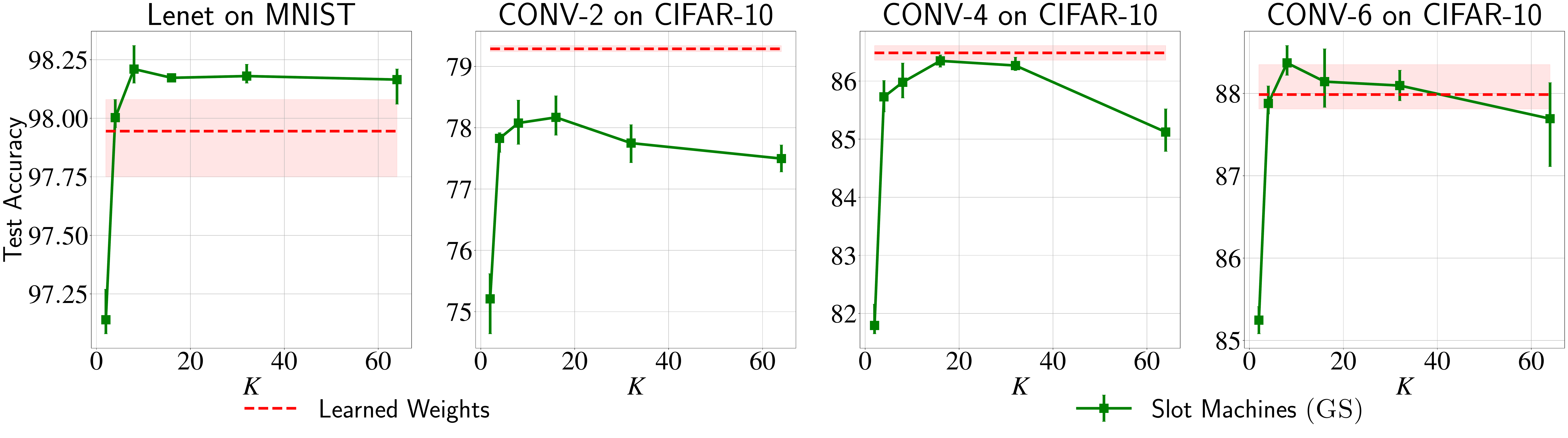}
    \caption{\textbf{Comparison with traditional training on CIFAR-10 and MNIST. } Performance of slot machines improves as K increases (here we consider $K \in \{2, 4, 8, 16, 32, 64\}$) although the performance degrades after $K\geq 8$ are small. For CONV-6 (the deepest model considered here), our approach using $\textsc{GS}$ achieves accuracy superior to that obtained with trained weights, while for CONV-4 it produces performance only slightly inferior to that of the optimized network. Furthermore, as illustrated by the error bars in these plots, the accuracy variances of slot machines are much smaller than those of networks traditionally trained by optimizing weights. Accuracies are measured on the {\em test} set over four different trials using early stopping on the {\em validation} accuracy. }
    \label{fig:k-compare-cifar10}
\end{figure*}

\begin{figure}[t]
 \centering
    \includegraphics[width=\linewidth]{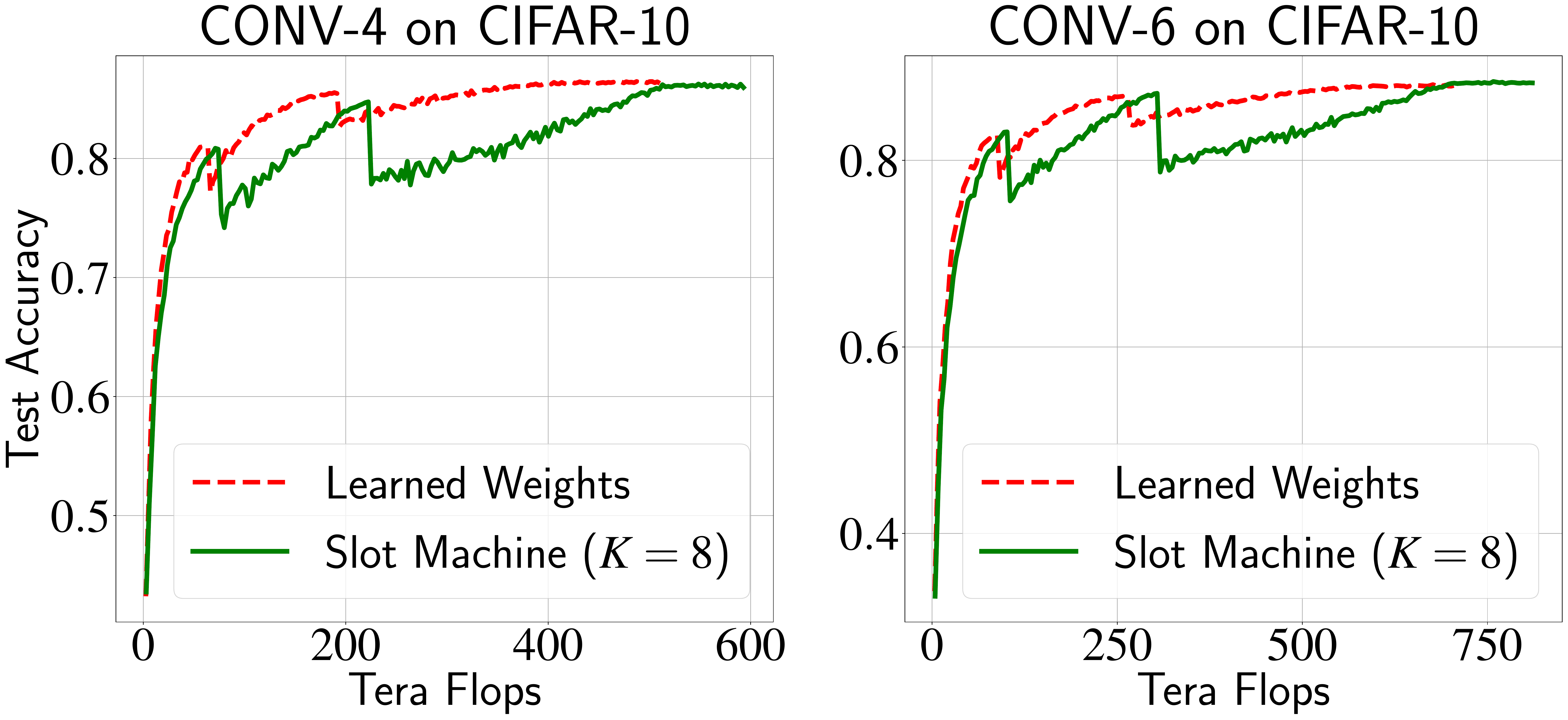}
    \caption{\textbf{Test accuracy versus training cost.} For the same training compute budget, Slot Machines achieve comparable performance to models traditionally optimized.}
    \label{fig:acc-flops}
\end{figure}

All models use a batch size of $128$ and stochastic gradient descent with warm restarts\footnote{Restarts happen at epoch 25 and 75.} \citep{loshchilov-ICLR17SGDR}, a momentum of 0.9 and an $\ell_2$ penalty of $0.0001$\footnote{\textsc{PS} models do not use weight-decay.}.  When training $\textsc{GS}$ slot machines, we set the learning rate to $0.2$ for $K \leq 8$ and $0.1$ otherwise. We set the learning rate to $0.01$ when directly optimizing the weights (training from scratch and finetuning) except when training VGG-19 where we set set the learning rate to $0.1$. We find that a high learning rate is required when sampling the network probabilistically, a behaviour which was also observed in~\citet{zhou2019deconstructing}. Accordingly, we use a learning rate of $25$ for all $\textsc{PS}$ models. We did not train VGG-19 using PS.

We use data augmentation and dropout (with a rate of $p = 0.5$) when experimenting on CIFAR-10~\citep{Krizhevsky09learningmultiple}.  We use batch normalization in VGG-19 but the affine parameters are never updated throughout training. 


\subsection{Slot Machines versus Traditionally-Trained Networks}\label{comparison-with-trained}
We compare the networks using random weights selected from our approach with two different baselines: (1) randomly initialized networks with one weight option per connection, and (2) traditionally-trained networks whose continuous weights are iteratively updated. These baselines are off-the-shelf modules from PyTorch~\citep{paszke2019pytorch} which we train in the standard way as explained in Section~\ref{setup}. For this first set of experiments we use GS to optimize Slot Machines, since it tends to provide better performance than PS (the two methods will be compared in subsection~\ref{selection-method}).

As shown in Figure~\ref{fig:comparision-random}, untrained networks with only one random weight per edge perform at chance. However, methodologically selecting the parameters from just two random values for each connection greatly enhances performance across different datasets and networks.  Even better, as shown in Figure~\ref{fig:k-compare-cifar10}, as the number of random weight options per connection increases, the performance of these networks approaches that of traditionally-trained networks with the same number of parameters, despite containing only random values. \cite{malach2020proving} proved that any ``ReLU network of depth $\ell$ can be approximated by finding a weighted-subnetwork of a random network of depth $2\ell$ and sufficient width.'' Without pruning, our selection method finds within the superset of fixed random weights a $6$ layer configuration that outperforms a $6$ layer traditionally-trained network. Furthermore, Figure~\ref{fig:acc-flops} shows that the overall cost of training Slot Machines is comparable to that of traditional optimization.

\begin{figure}[t]
    \centering
    \includegraphics[height=0.48\linewidth]{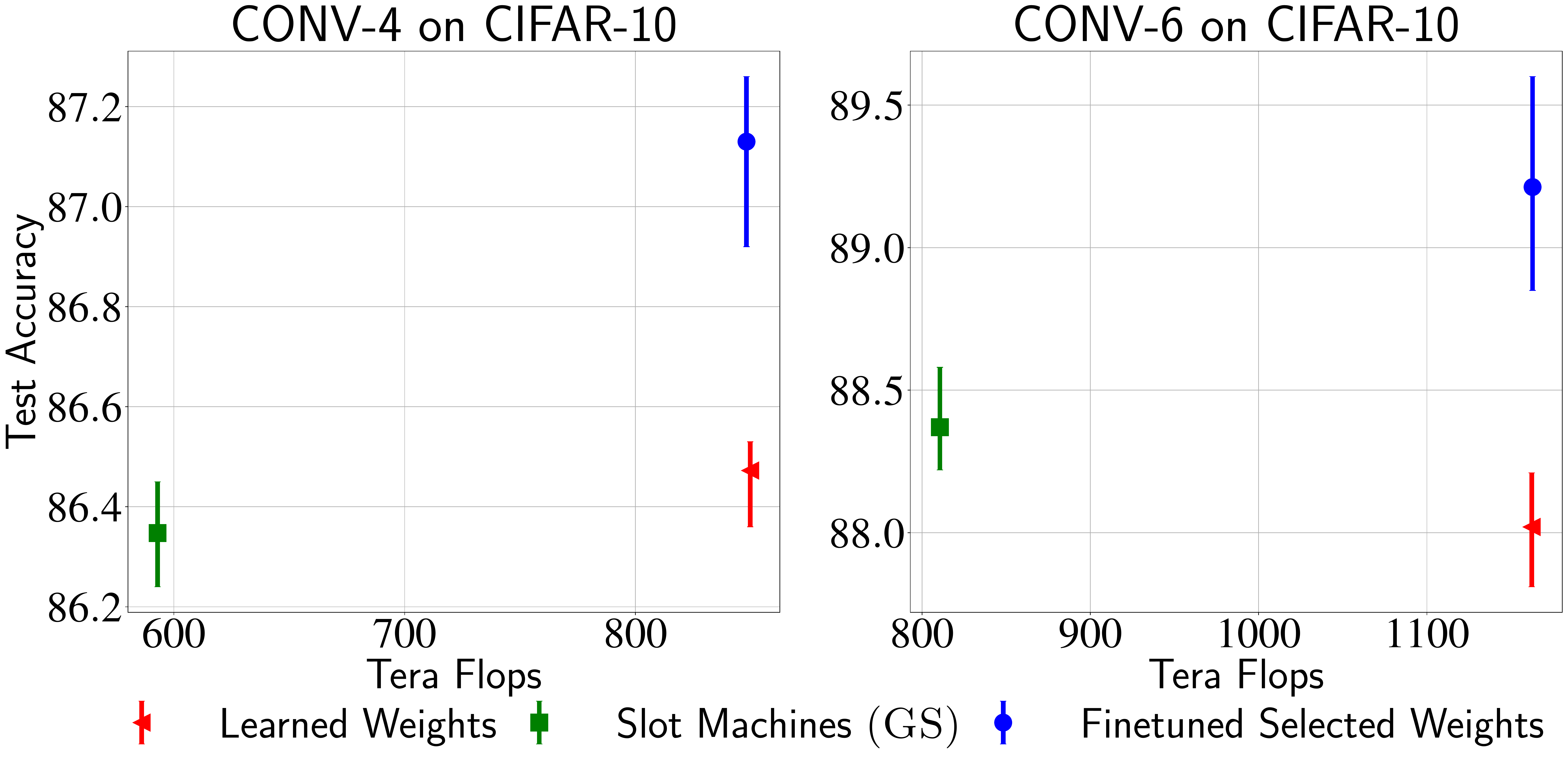}
    \caption{\textbf{Finetuning Slot Machines.} For the {\em same total training cost} (shown on the horizontal axis), CONV-4 and CONV-6 Slot Machines finetuned with traditional optimization achieve {\em better accuracy} compared to the same networks learned from scratch.}
    \label{fig:fine-tuning}
\end{figure}

\subsection{Finetuning Slot Machines}\label{finetuning}
Our approach can also be viewed as a strategy to provide a better initialization for traditional training. To assess the value of such a scheme, we finetune the networks obtained after training slot machines for $100$ epochs to match the cost of learned weights. 
Figure~\ref{fig:fine-tuning} summarizes the results in terms of training time (including both selection and finetuning) vs test accuracy. It can be noted that for the CONV-4 and CONV-6 architectures, finetuned slot machines achieve higher accuracy compared to the same models learned from scratch, at no additional training cost. For VGG-19, finetuning improves accuracy ($92.1\%$ instead of $91.7\%$) but the resulting model still does not match the performance of the model trained from scratch ($92.6\%$).

\begin{figure}[t]
    \centering
    \includegraphics[height=0.48\linewidth]{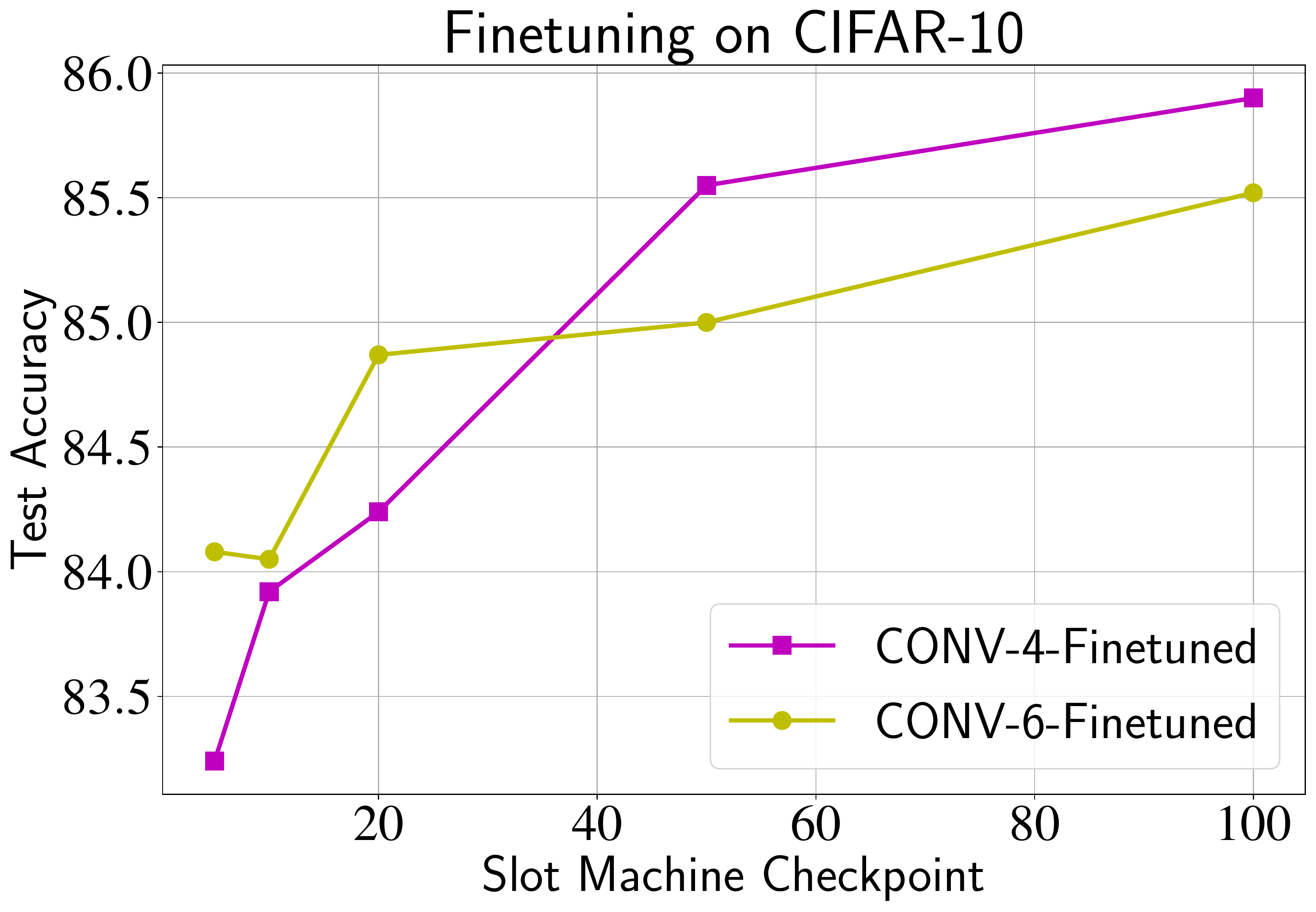}
    \caption{\textbf{Finetuning from different Slot Machine checkpoints.} Slot Machine checkpoint shows the number of training epochs used for weight selection before switching to finetuning (performed for 100 epochs). Performance is measured on the test set using early stopping determined by the maximum validation accuracy during finetuning. 
    }
    \label{fig:fine-tuning-checkpoint}
\end{figure}

To show that the weight selection in slot machines does in fact impact performance of finetuned models, we finetune from different slot machine checkpoints. If the selection is beneficial, then finetuning from later checkpoints will show improved performance. As shown in Figure~\ref{fig:fine-tuning-checkpoint}, this is indeed the case as finetuning from later checkpoints results in higher performance on the test set.

\subsection{Greedy Selection Versus Probabilistic Sampling}\label{selection-method}
As detailed in Section~\ref{main-algorithm}, we consider two different methods for sampling our networks in the forward pass: a greedy selection where the weight corresponding to the highest score is used and a stochastic selection which draws from a proper distribution over the weights.

\begin{figure}[!t]
\centering
\includegraphics[height=0.48\linewidth]{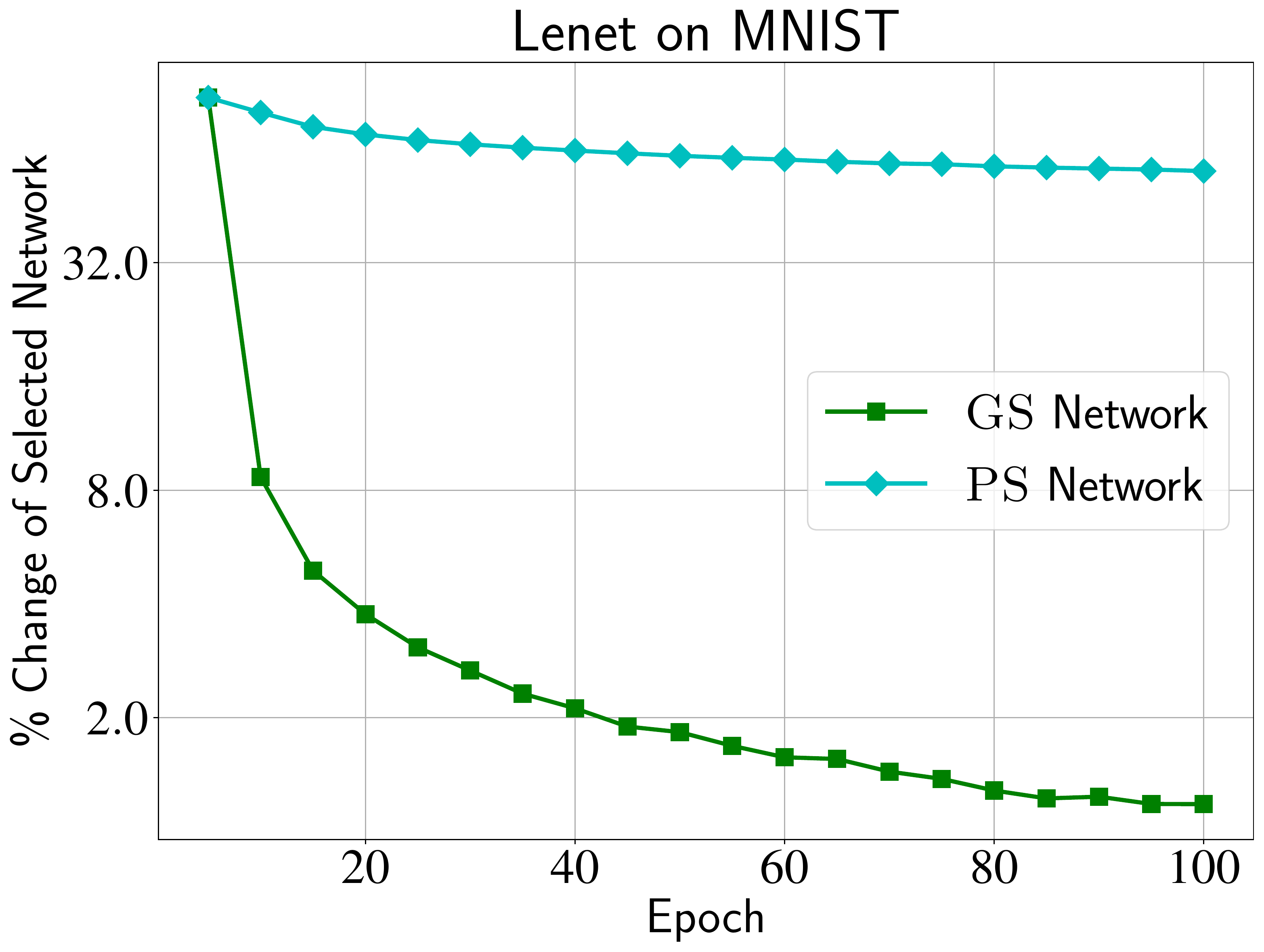}
      \caption{\textbf{Weight exploration in Slot Machines.} The vertical axis shows (on a log scale) the percentage of weights changed after every five epochs as training progresses.  Compared to PS, GS is much less exploratory and converges rapidly to a preferred configuration of weights. On the other hand, due to its probabilistic sampling, PS keeps changing the weight selections even in late stages of training. }
      \label{fig:subnet-change}
\end{figure}

To fully comprehend the behavior differences between these two strategies, it is instructive to look at Figure~\ref{fig:subnet-change}, which reports the percentage of weights changed every 5 epochs by the two strategies. PS keeps changing a large percentage of weights even in late stages of the optimization, due to its probabilistic sampling.

\begin{figure*}[!t]
    \centering
    \includegraphics[width=0.9\linewidth]{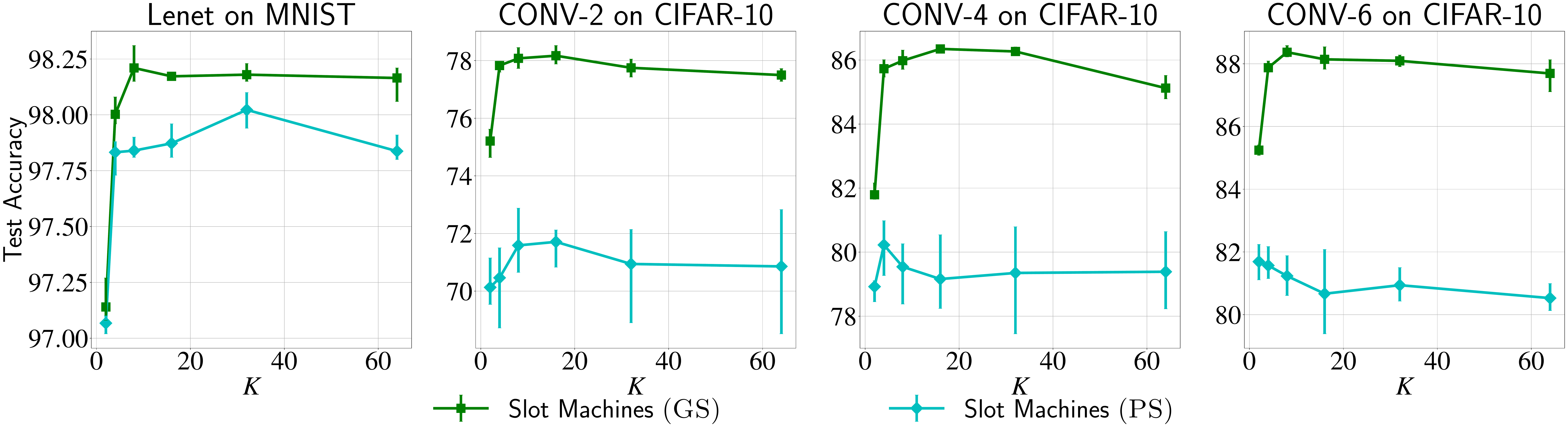}
    \caption{\textbf{Performance of GS vs PS.} Training slot machines via greedy selection (GS) yields better accuracy than optimizing them with probabilistic sampling (PS) for all values of $K$ considered. The reason is that $\textsc{PS}$ is a lot more exploratory and tends to produce slower convergence, as shown in Figure~\ref{fig:subnet-change}.}
    \label{fig:greedy-prob-compare}
\end{figure*}

As seen in Figure~\ref{fig:greedy-prob-compare}, \textsc{GS} performs better than \textsc{PS}. Despite the network changing considerably, \textsc{PS} still manages to obtain decent accuracy indicating that there are potentially many good random networks within a slot machine. However, as hypothesized in~\citet{ramanujan2019whats}, the high variability due to stochastic sampling means that the same network is likely never or rarely observed more than once in any training run. This makes learning extremely challenging and consequently adversely impacts performance. Conversely, \textsc{GS} is less exploratory and converges fairly quickly to a stable set of weights. 

From Figure~\ref{fig:greedy-prob-compare} we can also notice that the accuracy of \textsc{GS} improves or remains stable as the value of $K$ is increased. This is not always the case for \textsc{PS} when $K \geq 8$. We claim this behavior is expected since \textsc{GS} is more restricted in terms of the choices it can take. Thus, \textsc{GS} benefits more from large values of $K$ compared to \textsc{PS}.

\subsection{Sharing Random Weights}
Inspired by quantized networks~\citep{CourbariauxB16, RastegariORF16, hubara2017quantized, Wang2018}, we consider slot machines under two new settings. The first constrains the connections in a layer to share the same set of $K$ random weights. The second setting is even more restricting as it requires all connections in the network to share the same set of $K$ random weights. Under the first setting, at each layer the weights are drawn from the uniform distribution $(-\sigma_\ell, \sigma_\ell)$ where $\sigma_\ell$ is the standard deviation of the Glorot Normal distribution for layer $\ell$. When using a single set of weights for the entire network, we sample the weights independently from $\mathbb{U}(-\hat \sigma, \hat \sigma)$. $\hat \sigma$ is the mean of the standard deviations of the per layer Glorot Normal distributions. 

Each of the weights is still associated with a score. The slot machine with shared weights is then trained as before. This approach has the potential of compressing the model although the full set of of scores is still needed. 

\begin{figure}
    \centering
    \includegraphics[width=\linewidth]{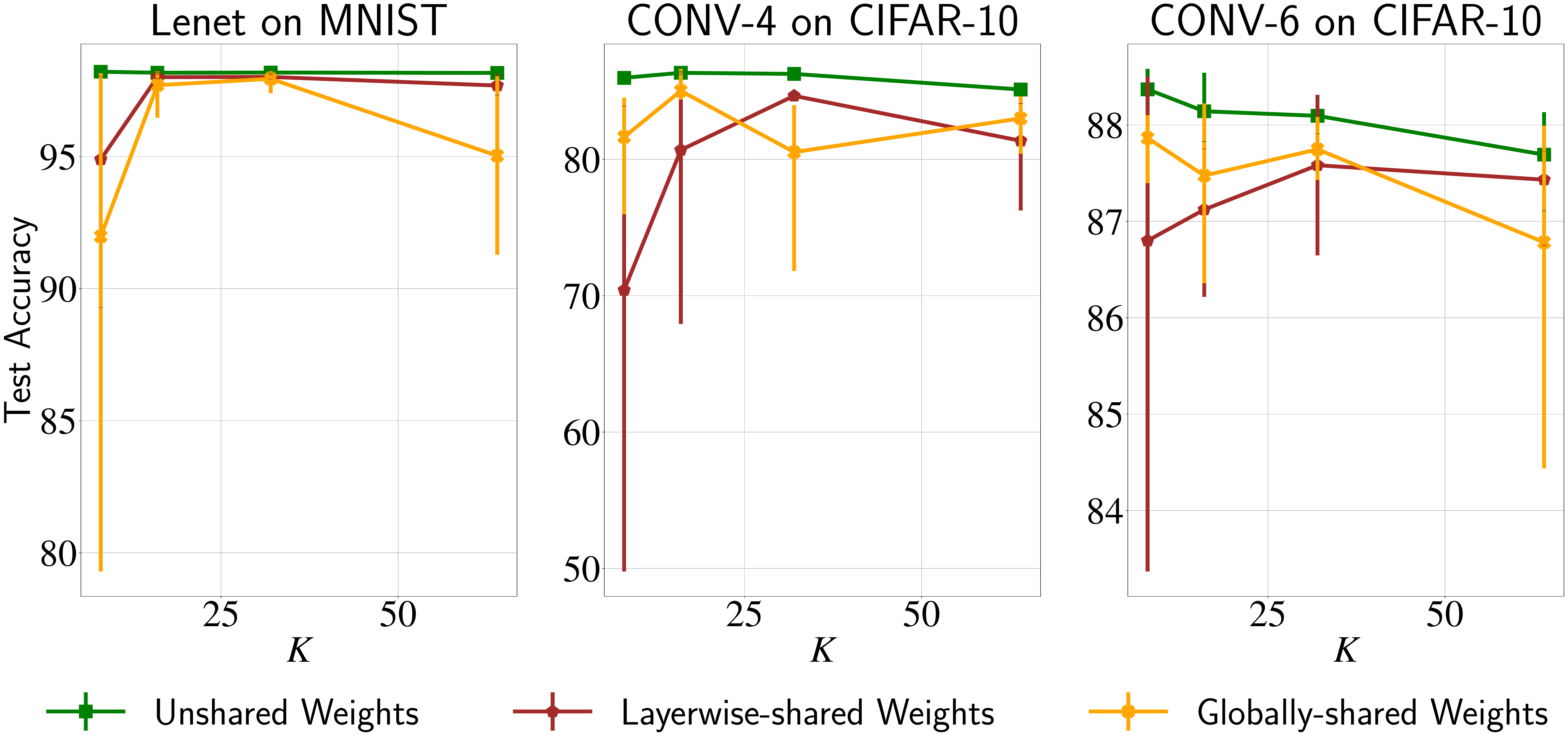}
    \caption{\textbf{Sharing random weights.} Slot Machines (\textsc{GS}) using the same set of $K$ random weights for all connections in a layer or even in the entire network perform quite well. However, they do not match the performance of Slot Machines (\textsc{GS}) that use different sets of weights for different connections. The benefit of sharing weights is that the memory requirements and space needed to store these networks is substantially smaller compared to the storage footprint of slot machines with unshared weights. As an example, a Lenet model with unshared weights needs $\sim 1$MB of storage whereas the same model using shared weights in each layer needs $\sim0.02$MB of storage.} 
    \label{fig:shared-weights}
\end{figure}

As shown in Figure~\ref{fig:shared-weights}, these models continue to do well when $K$ is large enough. However, unlike conventional slot machines, these models do not work when $K$ is very small, e.g., $K = 2$. Furthermore, the accuracy exhibits a large variance from run to run, as evidenced by the large error bars in the plot. This is understandable, as the slot machine with shared weights is restricted to search in a much smaller space of parameter combinations and thus the probability of finding a winning combination is much reduced.

\textbf{Difference between Slot Machines (SMs) and quantized networks. } Although in the experiment above we evaluate SMs under the shared-weight setting commonly adopted in network quantization, we would like to point out that several differences separate our approach from prior work in network quantization. (1) {\em Goal.} The aim of network quantization is model compression and increased efficiency. Conversely, the goal of SMs is to achieve accuracy comparable or better than traditional training by means of weight selection instead of continuous optimization. We also demonstrate that finetuning SMs via continuous optimization results in higher accuracy compared to training from scratch as shown  Figure~\ref{fig:fine-tuning}. (2) {\em Shared vs unshared weights.} Quantized networks use shared weights across all connections. For example, BinaryNets~\citep{CourbariauxB16} use only weight values $+1/-1$. While this makes sense for the purpose of reducing the model footprint and increasing efficiency, sharing weights causes a drop in accuracy (e.g., compare performance of Globally-shared vs Unshared in Figure ~\ref{fig:shared-weights}). SMs use distinct sets of weights for different connections (i.e., unshared weights) in order to retain high accuracy. (3) {\em Optimization.} While quantization networks typically involve a continuous optimization over the weights, our approach involves discrete selection of one out of $K$ fixed weights for each connection.

    \subsection{Sparse Slot Machines } 
We conducted experiments where one of the $K$ weight options is constrained to always be $0$ which induced sparse networks. However, the resulting sparsity is low when $K$ is large. For CONV-6 on CIFAR-10, the sparsity is $49\%$ when $K=2$, and $1.1\%$ when $K=64$.  If  $K$ is small, the selected sparse network has a lower performance compared to the corresponding standard Slot Machine where all the $K$ weight options are initialized randomly  (76\% versus 83\% test accuracy for CONV-6 on CIFAR-10 with  $K=2$). However, when $K$ is large (e.g., $K=64$), the sparse network has performance comparable to  a standard Slot Machine. This is because when $K$ is small, restricting one of the weights to be $0$ effectively removes a possible non-zero value from the already few options.

\subsection{Experimental Comparison with Prior Pruning Approaches}
\begin{figure}
    \centering
    \begin{subfigure}{0.48\linewidth}
    \includegraphics[width=0.9\linewidth]{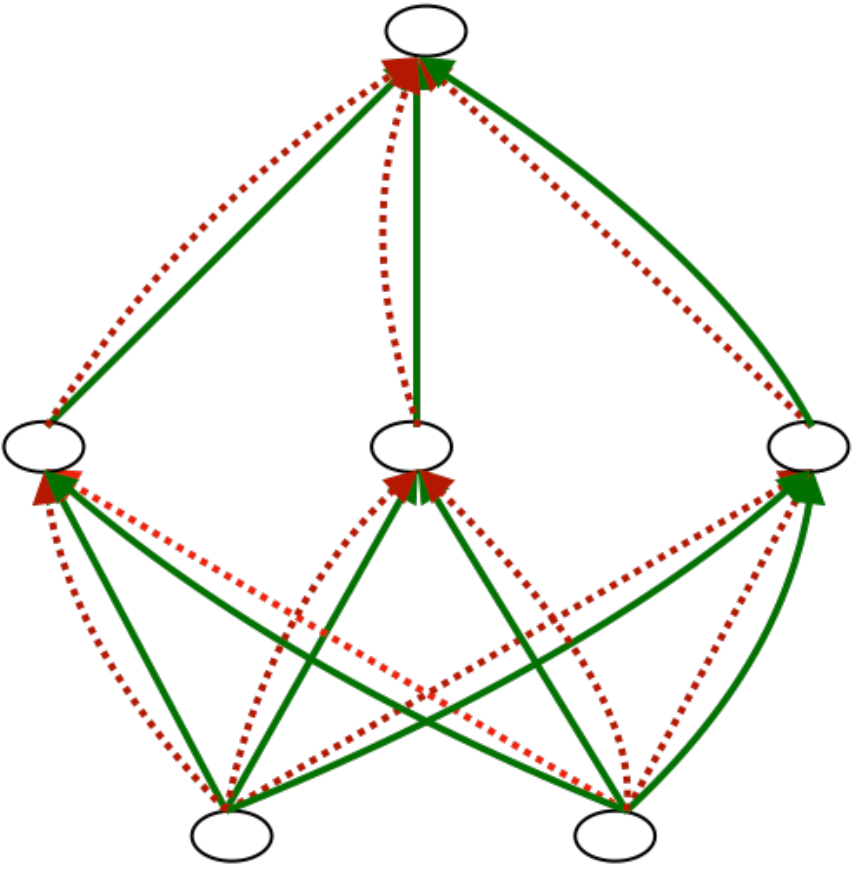}
    \end{subfigure}
    \begin{subfigure}{0.48\linewidth}
    \includegraphics[width=0.9\linewidth]{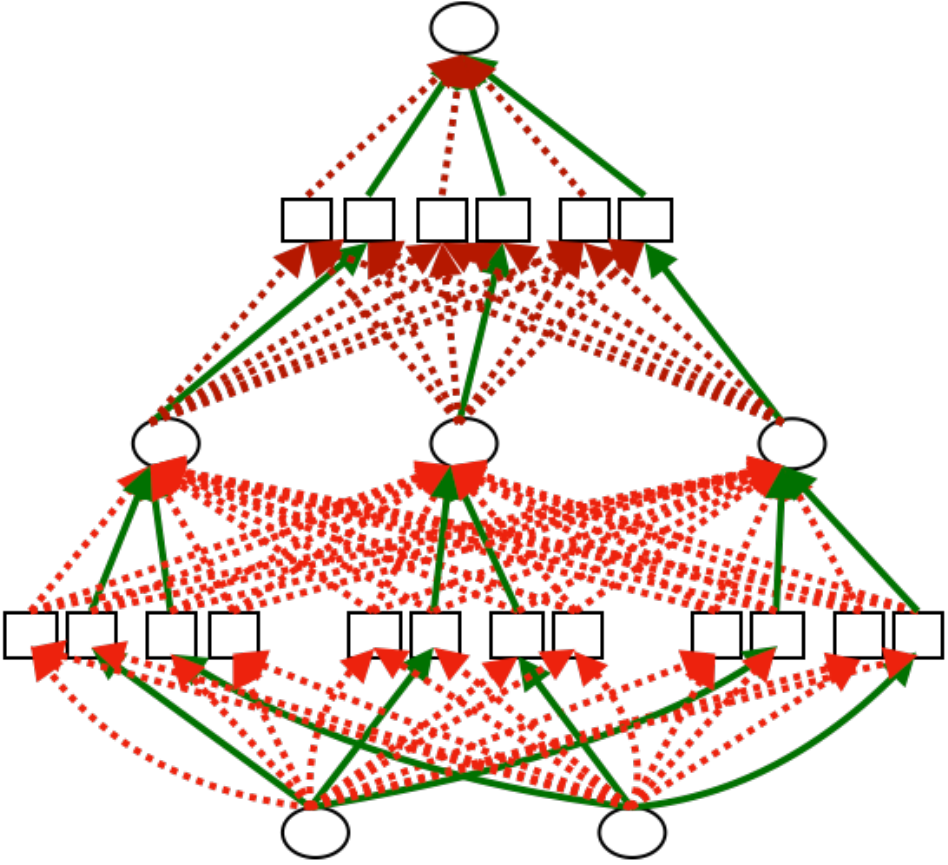}
    \end{subfigure}
    \caption{In principle, it is possible to obtain our network (\emph{left}) by pruning a bigger network constructed ad-hoc (\emph{right}). In this example, our slot machine uses $K = 2$ options per connection ($i, j$). The green-colored connections represent the selected weights. The square boxes in the bigger network implement identity functions. The circles designate vanilla neurons with non-linear activation functions. Red dash lines in our network represent unchosen weights. These lines in the bigger network would correspond to pruned weights.} 
    \label{fig:relation-to-pruning}
\end{figure}

The models learned by our algorithm could in principle be found by applying pruning to a bigger network representing the multiple weight options in the form of additional connections. One way to achieve this is by introducing additional ``dummy" layers after every layer $\ell$ except the output layer. Each ``dummy" layer will have $K * c$ identity units where $c = n_{\ell} * n_{\ell + 1}$ and $n_{\ell}$ is the number of neurons in layer $\ell$. The addition of these layers has the effect of separating out the $K$ random values for each connection in our network into distinct connection weights. It is important that the neurons of the ``dummy" layer encode the identity function to ensure that the random values can pass through it unmodified. Finally, in order to obtain the model learned by our system, all connections between a layer and its associated ``dummy" layer must be pruned except for the weights which would have been selected by our algorithm as shown in Figure~\ref{fig:relation-to-pruning}. This procedure requires allocating a bigger network and is clearly more costly compared to our algorithm. 

In this section, we compare slot machines with pruning techniques in prior works. Our approach is similar to the pruning technique of~\citet{ramanujan2019whats} as their method too does not update the weights of the networks after initialization. Furthermore, their strategy selects the weights greedily, as in our $\textsc{GS}$. However, they use one weight per connection and employ pruning to uncover good subnetworks within the random network whereas we use multiple random values per edge. Additionally, we do not ever prune any of the connections. We compare the results of our networks to this prior work in Table~\ref{tab:comparison-with-pruning}. We also compare with supermasks~\citep{zhou2019deconstructing}. Supermasks employ a probability distribution during the selection which makes them reminiscent of our $\textsc{PS}$ models. However, they use a Bernoulli distribution at each weight while $\textsc{PS}$ uses a Multinomial distribution at each connection. Also, like ~\citet{ramanujan2019whats}, supermasks have one weight per connection and perform pruning rather than weight selection. Table~\ref{tab:comparison-with-pruning} shows that $\textsc{GS}$ achieves accuracy comparable to that of~\citet{ramanujan2019whats} while $\textsc{PS}$ matches the performance of supermasks. These results suggest an interesting empirical performance equivalency among these related but distinct approaches. 

\begin{table}
\caption{Comparison with~\cite{ramanujan2019whats} and~\citep{zhou2019deconstructing}. The results of the first two rows are from the respective papers.}
    \label{tab:comparison-with-pruning}
    \centering
    \vskip 0.15in
    \resizebox{\linewidth}{!}{\begin{tabular}{lcccc}
    \toprule
        Method & Lenet & CONV-2 & CONV-4 & CONV-6\\
        \midrule
        \citet{ramanujan2019whats} & - & 77.7 & 85.8 & 88.1 \\
        \midrule
        SNIP~\citep{zhou2019deconstructing} & 98.0 & 66.0 & 72.5 & 76.5 \\
        \midrule
        \midrule
        Slot Machines ($\textsc{GS}$)  & 98.2 & 78.2 & 86.3 & 88.4 \\
        \midrule
         Slot Machines ($\textsc{PS}$) & 98.0 & 71.7 & 80.2 & 81.7  \\
        \bottomrule
    \end{tabular}}
    \vskip 0.1in
\end{table}

\subsection{Distribution of Selected Weights}\label{distributions-weights}

\begin{figure}
    \centering
    \includegraphics[width=\linewidth]{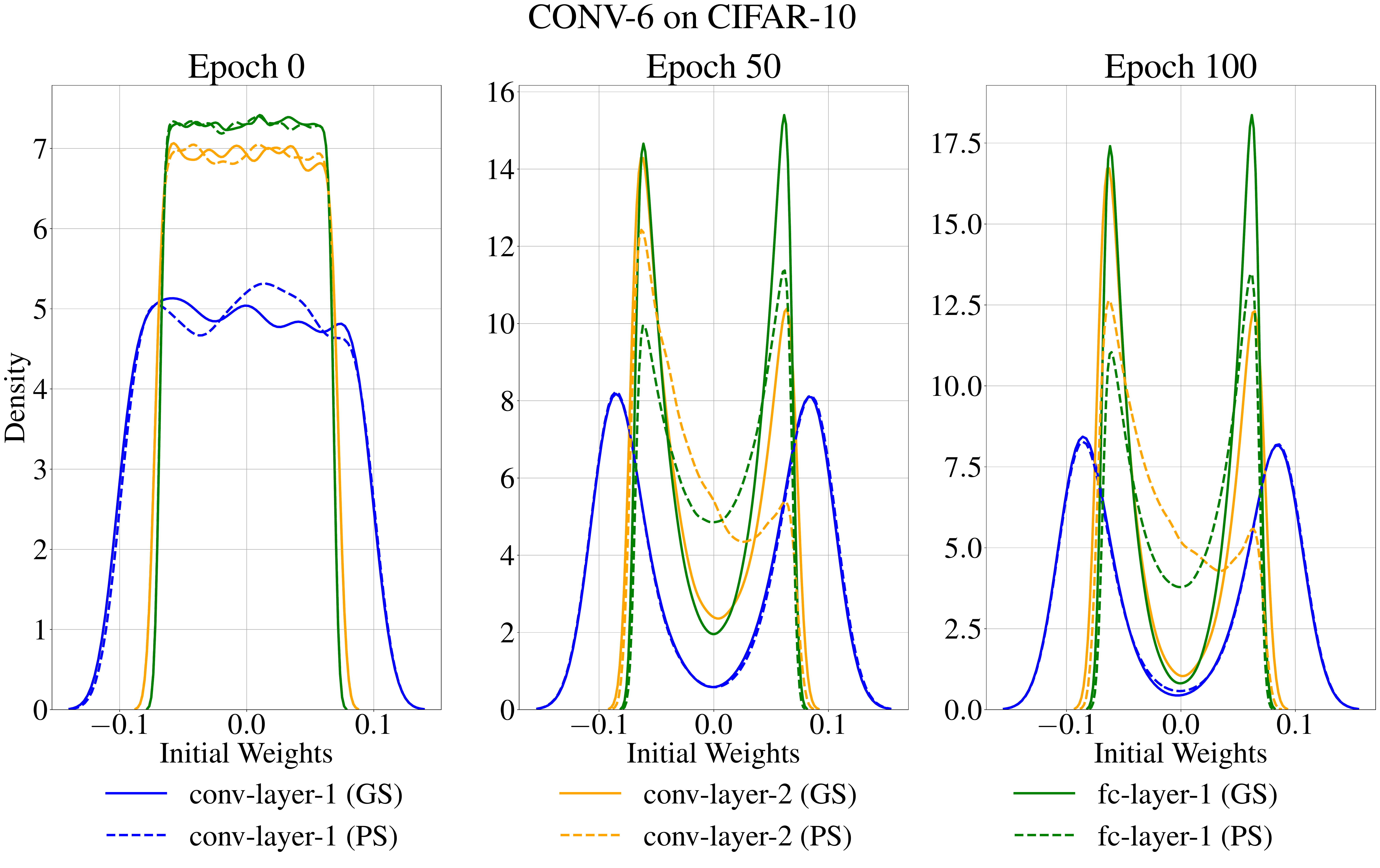}
    \caption{The distributions of the selected weights in the first two convolutional and the first fully-connected layers of CONV-6 on CIFAR-10. Starting from purely uniform distributions, Slot Machines progressively choose large magnitude weights as training proceeds. See supplementary material for additional analyses.}
    \label{fig:weight-distribution-conv6}
\end{figure}

In Figure~\ref{fig:weight-distribution-conv6} we study the distribution of selected weights at different training points in order to understand why certain weights are chosen and others are not.  We observe that both $\textsc{GS}$ and $\textsc{PS}$ tend to prefer weights having large magnitudes as learning progresses.  This propensity for large weights might help explain why methods such as magnitude-based pruning of traditionally-trained networks work as well as they do. We provide further analyses of this phenomenon in the supplementary material. 


\section{Conclusion and Future Work}

This work shows that neural networks with random weights perform competitively, provided that each connection is given multiple weight options and that a good selection strategy is used. We introduce a simple selection procedure that is remarkably effective and consistent in producing strong weight configurations from  few random options per connection. We also demonstrate that these selected configurations can be used as starting initializations for finetuning, which often produces accuracy gains over training the network from scratch, at comparable computational cost. Our study suggests that our method tends to naturally select large magnitude weights as training proceeds. Future work will be devoted to further analyze what other properties differentiate selected weights  from those that are not selected, as knowing such properties may pave the way for more effective initializations for neural networks. More work is also needed to reduce the memory requirements of these networks so they can be scaled to bigger networks.

\bibliography{slot-machines}

\begin{thebibliography}{36}
\providecommand{\natexlab}[1]{#1}
\providecommand{\url}[1]{\texttt{#1}}
\expandafter\ifx\csname urlstyle\endcsname\relax
  \providecommand{\doi}[1]{doi: #1}\else
  \providecommand{\doi}{doi: \begingroup \urlstyle{rm}\Url}\fi

\bibitem[Bengio et~al.(2013)Bengio, Léonard, and
  Courville]{bengio2013estimating}
Bengio, Y., Léonard, N., and Courville, A.
\newblock Estimating or propagating gradients through stochastic neurons for
  conditional computation, 2013.

\bibitem[Breiman et~al.(1984)Breiman, Friendman, Stone, and Olstein]{Breiman84}
Breiman, L., Friendman, J., Stone, C.~J., and Olstein, R.~A.
\newblock \emph{Classification and regression trees.}
\newblock Wadsworth \& Brooks/Cole Advanced Books \& Software., Monterey, CA,
  1984.
\newblock ISBN 978-0-412-04841-8.

\bibitem[Brown et~al.(2020)Brown, Mann, Ryder, Subbiah, Kaplan, Dhariwal,
  Neelakantan, Shyam, Sastry, Askell, Agarwal, Herbert-Voss, Krueger, Henighan,
  Child, Ramesh, Ziegler, Wu, Winter, Hesse, Chen, Sigler, Litwin, Gray, Chess,
  Clark, Berner, McCandlish, Radford, Sutskever, and Amodei]{brown2020language}
Brown, T.~B., Mann, B., Ryder, N., Subbiah, M., Kaplan, J., Dhariwal, P.,
  Neelakantan, A., Shyam, P., Sastry, G., Askell, A., Agarwal, S.,
  Herbert-Voss, A., Krueger, G., Henighan, T., Child, R., Ramesh, A., Ziegler,
  D.~M., Wu, J., Winter, C., Hesse, C., Chen, M., Sigler, E., Litwin, M., Gray,
  S., Chess, B., Clark, J., Berner, C., McCandlish, S., Radford, A., Sutskever,
  I., and Amodei, D.
\newblock Language models are few-shot learners.
\newblock In \emph{Advances in Neural Information Processing Systems},
  volume~33, 2020.
\newblock URL
  \url{https://papers.nips.cc/paper/2020/file/1457c0d6bfcb4967418bfb8ac142f64a-Paper.pdf}.

\bibitem[Frankle \& Carbin(2019)Frankle and Carbin]{frankle2018lottery}
Frankle, J. and Carbin, M.
\newblock The lottery ticket hypothesis: Finding sparse, trainable neural
  networks.
\newblock In \emph{International Conference on Learning Representations}, 2019.
\newblock URL \url{https://openreview.net/forum?id=rJl-b3RcF7}.

\bibitem[Gaier \& Ha(2019)Gaier and Ha]{gaier2019weight}
Gaier, A. and Ha, D.
\newblock Weight agnostic neural networks.
\newblock In Wallach, H., Larochelle, H., Beygelzimer, A., d\textquotesingle
  Alch\'{e}-Buc, F., Fox, E., and Garnett, R. (eds.), \emph{Advances in Neural
  Information Processing Systems}, volume~32, pp.\  5364--5378. Curran
  Associates, Inc., 2019.
\newblock URL
  \url{https://proceedings.neurips.cc/paper/2019/file/e98741479a7b998f88b8f8c9f0b6b6f1-Paper.pdf}.

\bibitem[Glorot \& Bengio(2010)Glorot and Bengio]{pmlr-v9-glorot10a}
Glorot, X. and Bengio, Y.
\newblock Understanding the difficulty of training deep feedforward neural
  networks.
\newblock In Teh, Y.~W. and Titterington, M. (eds.), \emph{Proceedings of the
  Thirteenth International Conference on Artificial Intelligence and
  Statistics}, volume~9 of \emph{Proceedings of Machine Learning Research},
  pp.\  249--256, Chia Laguna Resort, Sardinia, Italy, 13--15 May 2010. PMLR.

\bibitem[{He} et~al.(2015){He}, {Zhang}, {Ren}, and {Sun}]{He2050}
{He}, K., {Zhang}, X., {Ren}, S., and {Sun}, J.
\newblock Delving deep into rectifiers: Surpassing human-level performance on
  imagenet classification.
\newblock In \emph{2015 IEEE International Conference on Computer Vision
  (ICCV)}, pp.\  1026--1034, 2015.

\bibitem[He et~al.(2016)He, Zhang, Ren, and Sun]{resnet-34}
He, K., Zhang, X., Ren, S., and Sun, J.
\newblock Deep residual learning for image recognition.
\newblock In \emph{2016 IEEE Conference on Computer Vision and Pattern
  Recognition (CVPR)}, pp.\  770--778, 2016.

\bibitem[He et~al.(2017)He, Gkioxari, Dollár, and Girshick]{he2017mask}
He, K., Gkioxari, G., Dollár, P., and Girshick, R.
\newblock Mask r-cnn.
\newblock In \emph{2017 IEEE International Conference on Computer Vision
  (ICCV)}, pp.\  2980--2988, 2017.
\newblock \doi{10.1109/ICCV.2017.322}.

\bibitem[Hoffer et~al.(2018)Hoffer, Hubara, and Soudry]{hoffer2018fix}
Hoffer, E., Hubara, I., and Soudry, D.
\newblock Fix your classifier: the marginal value of training the last weight
  layer.
\newblock In \emph{International Conference on Learning Representations}, 2018.
\newblock URL \url{https://openreview.net/forum?id=S1Dh8Tg0-}.

\bibitem[Hubara et~al.(2016)Hubara, Courbariaux, Soudry, El-Yaniv, and
  Bengio]{CourbariauxB16}
Hubara, I., Courbariaux, M., Soudry, D., El-Yaniv, R., and Bengio, Y.
\newblock Binarized neural networks.
\newblock In Lee, D., Sugiyama, M., Luxburg, U., Guyon, I., and Garnett, R.
  (eds.), \emph{Advances in Neural Information Processing Systems}, volume~29.
  Curran Associates, Inc., 2016.
\newblock URL
  \url{https://proceedings.neurips.cc/paper/2016/file/d8330f857a17c53d217014ee776bfd50-Paper.pdf}.

\bibitem[Hubara et~al.(2017)Hubara, Courbariaux, Soudry, El-Yaniv, and
  Bengio]{hubara2017quantized}
Hubara, I., Courbariaux, M., Soudry, D., El-Yaniv, R., and Bengio, Y.
\newblock Quantized neural networks: Training neural networks with low
  precision weights and activations.
\newblock \emph{The Journal of Machine Learning Research}, 18\penalty0
  (1):\penalty0 6869--6898, 2017.

\bibitem[Krizhevsky(2009)]{Krizhevsky09learningmultiple}
Krizhevsky, A.
\newblock Learning multiple layers of features from tiny images.
\newblock Technical report, University of Toronto, 2009.

\bibitem[Krizhevsky et~al.(2012)Krizhevsky, Sutskever, and
  Hinton]{Krizhevsky-imagenetclassification}
Krizhevsky, A., Sutskever, I., and Hinton, G.~E.
\newblock Imagenet classification with deep convolutional neural networks.
\newblock In Pereira, F., Burges, C. J.~C., Bottou, L., and Weinberger, K.~Q.
  (eds.), \emph{Advances in Neural Information Processing Systems}, volume~25,
  pp.\  1097--1105. Curran Associates, Inc., 2012.
\newblock URL
  \url{https://proceedings.neurips.cc/paper/2012/file/c399862d3b9d6b76c8436e924a68c45b-Paper.pdf}.

\bibitem[Lecun et~al.(1998)Lecun, Bottou, Bengio, and Haffner]{lecun1998lenet}
Lecun, Y., Bottou, L., Bengio, Y., and Haffner, P.
\newblock Gradient-based learning applied to document recognition.
\newblock \emph{Proceedings of the IEEE}, 86\penalty0 (11):\penalty0
  2278--2324, 1998.

\bibitem[Lee et~al.(2019)Lee, Ajanthan, and Torr]{lee2018snip}
Lee, N., Ajanthan, T., and Torr, P.
\newblock {SNIP}: {Single}-{shot} {Network} {Pruning} {based} {on} {connection}
  {sensitivity}.
\newblock In \emph{International Conference on Learning Representations}, 2019.
\newblock URL \url{https://openreview.net/forum?id=B1VZqjAcYX}.

\bibitem[Lee et~al.(2020)Lee, Ajanthan, Gould, and Torr]{lee2019signal}
Lee, N., Ajanthan, T., Gould, S., and Torr, P. H.~S.
\newblock A signal propagation perspective for pruning neural networks at
  initialization.
\newblock In \emph{International Conference on Learning Representations}, 2020.
\newblock URL \url{https://openreview.net/forum?id=HJeTo2VFwH}.

\bibitem[Loshchilov \& Hutter(2017)Loshchilov and
  Hutter]{loshchilov-ICLR17SGDR}
Loshchilov, I. and Hutter, F.
\newblock Sgdr: Stochastic gradient descent with warm restarts.
\newblock In \emph{International Conference on Learning Representations
  (ICLR)}, 2017.
\newblock URL \url{https://openreview.net/pdf?id=Skq89Scxx}.

\bibitem[Maennel et~al.(2020)Maennel, Alabdulmohsin, Tolstikhin, Baldock,
  Bousquet, Gelly, and Keysers]{maennel2020neural}
Maennel, H., Alabdulmohsin, I., Tolstikhin, I., Baldock, R. J.~N., Bousquet,
  O., Gelly, S., and Keysers, D.
\newblock What do neural networks learn when trained with random labels?, 2020.
\newblock URL \url{https://arxiv.org/pdf/2006.10455.pdf}.

\bibitem[Malach et~al.(2020)Malach, Yehudai, Shalev-Schwartz, and
  Shamir]{malach2020proving}
Malach, E., Yehudai, G., Shalev-Schwartz, S., and Shamir, O.
\newblock Proving the lottery ticket hypothesis: Pruning is all you need.
\newblock In III, H.~D. and Singh, A. (eds.), \emph{Proceedings of the 37th
  International Conference on Machine Learning}, volume 119 of
  \emph{Proceedings of Machine Learning Research}, pp.\  6682--6691. PMLR,
  13--18 Jul 2020.
\newblock URL \url{http://proceedings.mlr.press/v119/malach20a.html}.

\bibitem[Paszke et~al.(2019)Paszke, Gross, Massa, Lerer, Bradbury, Chanan,
  Killeen, Lin, Gimelshein, Antiga, Desmaison, Kopf, Yang, DeVito, Raison,
  Tejani, Chilamkurthy, Steiner, Fang, Bai, and Chintala]{paszke2019pytorch}
Paszke, A., Gross, S., Massa, F., Lerer, A., Bradbury, J., Chanan, G., Killeen,
  T., Lin, Z., Gimelshein, N., Antiga, L., Desmaison, A., Kopf, A., Yang, E.,
  DeVito, Z., Raison, M., Tejani, A., Chilamkurthy, S., Steiner, B., Fang, L.,
  Bai, J., and Chintala, S.
\newblock Pytorch: An imperative style, high-performance deep learning library.
\newblock In Wallach, H., Larochelle, H., Beygelzimer, A., d\textquotesingle
  Alch\'{e}-Buc, F., Fox, E., and Garnett, R. (eds.), \emph{Advances in Neural
  Information Processing Systems}, volume~32, pp.\  8026--8037. Curran
  Associates, Inc., 2019.
\newblock URL
  \url{https://proceedings.neurips.cc/paper/2019/file/bdbca288fee7f92f2bfa9f7012727740-Paper.pdf}.

\bibitem[Pensia et~al.(2020)Pensia, Rajput, Nagle, Vishwakarma, and
  Papailiopoulos]{pensia2020optimal}
Pensia, A., Rajput, S., Nagle, A., Vishwakarma, H., and Papailiopoulos, D.
\newblock Optimal lottery tickets via subsetsum: Logarithmic
  over-parameterization is sufficient.
\newblock In \emph{Advances in Neural Information Processing Systems},
  volume~33, 2020.
\newblock URL
  \url{https://papers.nips.cc/paper/2020/file/1b742ae215adf18b75449c6e272fd92d-Paper.pdf}.

\bibitem[Ramanujan et~al.(2020)Ramanujan, Wortsman, Kembhavi, Farhadi, and
  Rastegari]{ramanujan2019whats}
Ramanujan, V., Wortsman, M., Kembhavi, A., Farhadi, A., and Rastegari, M.
\newblock What's hidden in a randomly weighted neural network?
\newblock In \emph{Proceedings of the IEEE/CVF Conference on Computer Vision
  and Pattern Recognition (CVPR)}, June 2020.
\newblock URL
  \url{https://openaccess.thecvf.com/content\_CVPR\_2020/papers/Ramanujan\_Whats\_Hidden\_in\_a\_Randomly\_Weighted_Neural\_Network\_CVPR\_2020\_paper.pdf}.

\bibitem[Rastegari et~al.(2016)Rastegari, Ordonez, Redmon, and
  Farhadi]{RastegariORF16}
Rastegari, M., Ordonez, V., Redmon, J., and Farhadi, A.
\newblock Xnor-net: Imagenet classification using binary convolutional neural
  networks.
\newblock In \emph{Computer Vision -- ECCV 2016}, pp.\  525--542, 2016.
\newblock URL \url{http://arxiv.org/abs/1603.05279}.

\bibitem[Ren et~al.(2015)Ren, He, Girshick, and Sun]{ren2015faster}
Ren, S., He, K., Girshick, R., and Sun, J.
\newblock Faster r-cnn: Towards real-time object detection with region proposal
  networks.
\newblock In Cortes, C., Lawrence, N., Lee, D., Sugiyama, M., and Garnett, R.
  (eds.), \emph{Advances in Neural Information Processing Systems}, volume~28,
  pp.\  91--99. Curran Associates, Inc., 2015.
\newblock URL
  \url{https://proceedings.neurips.cc/paper/2015/file/14bfa6bb14875e45bba028a21ed38046-Paper.pdf}.

\bibitem[Rosenfeld \& Tsotsos(2019)Rosenfeld and Tsotsos]{amir2019intri}
Rosenfeld, A. and Tsotsos, J.~K.
\newblock Intriguing properties of randomly weighted networks: Generalizing
  while learning next to nothing.
\newblock In \emph{2019 16th Conference on Computer and Robot Vision (CRV)},
  pp.\  9--16, 2019.

\bibitem[Russakovsky et~al.(2009)Russakovsky, Deng, Su, Krause, Satheesh, Ma,
  Huang, Karpathy, Khosla, Bernstein, Berg, and Fei-Fei]{imagenet-cvpr09}
Russakovsky, O., Deng, J., Su, H., Krause, J., Satheesh, S., Ma, S., Huang, Z.,
  Karpathy, A., Khosla, A., Bernstein, M., Berg, A.~C., and Fei-Fei, L.
\newblock {ImageNet: A Large-Scale Hierarchical Image Database}.
\newblock In \emph{CVPR}, 2009.

\bibitem[{Stanley} \& {Miikkulainen}(2002){Stanley} and
  {Miikkulainen}]{stanley2002neat}
{Stanley}, K.~O. and {Miikkulainen}, R.
\newblock Evolving neural networks through augmenting topologies.
\newblock \emph{Evolutionary Computation}, 10\penalty0 (2):\penalty0 99--127,
  2002.

\bibitem[Tanaka et~al.(2020)Tanaka, Kunin, Yamins, and
  Ganguli]{tanaka2020pruning}
Tanaka, H., Kunin, D., Yamins, D. L.~K., and Ganguli, S.
\newblock Pruning neural networks without any data by iteratively conserving
  synaptic flow.
\newblock In \emph{Advances in Neural Information Processing Systems},
  volume~33, 2020.
\newblock URL
  \url{https://proceedings.neurips.cc/paper/2020/file/46a4378f835dc8040c8057beb6a2da52-Paper.pdf}.

\bibitem[Vaswani et~al.(2017)Vaswani, Shazeer, Parmar, Uszkoreit, Jones, Gomez,
  Kaiser, and Polosukhin]{vaswani2017attention}
Vaswani, A., Shazeer, N., Parmar, N., Uszkoreit, J., Jones, L., Gomez, A.~N.,
  Kaiser, L.~u., and Polosukhin, I.
\newblock Attention is all you need.
\newblock In Guyon, I., Luxburg, U.~V., Bengio, S., Wallach, H., Fergus, R.,
  Vishwanathan, S., and Garnett, R. (eds.), \emph{Advances in Neural
  Information Processing Systems}, volume~30, pp.\  5998--6008. Curran
  Associates, Inc., 2017.
\newblock URL
  \url{https://proceedings.neurips.cc/paper/2017/file/3f5ee243547dee91fbd053c1c4a845aa-Paper.pdf}.

\bibitem[Wang et~al.(2020)Wang, Zhang, and Grosse]{wang2020picking}
Wang, C., Zhang, G., and Grosse, R.
\newblock Picking winning tickets before training by preserving gradient flow.
\newblock In \emph{International Conference on Learning Representations}, 2020.
\newblock URL \url{https://openreview.net/forum?id=SkgsACVKPH}.

\bibitem[{Wang} et~al.(2018){Wang}, {Hu}, {Zhang}, {Zhang}, {Liu}, and
  {Cheng}]{Wang2018}
{Wang}, P., {Hu}, Q., {Zhang}, Y., {Zhang}, C., {Liu}, Y., and {Cheng}, J.
\newblock Two-step quantization for low-bit neural networks.
\newblock In \emph{2018 IEEE/CVF Conference on Computer Vision and Pattern
  Recognition}, pp.\  4376--4384, 2018.
\newblock \doi{10.1109/CVPR.2018.00460}.

\bibitem[Yosinski et~al.(2014)Yosinski, Clune, Bengio, and
  Lipson]{yosinski2014transferable}
Yosinski, J., Clune, J., Bengio, Y., and Lipson, H.
\newblock How transferable are features in deep neural networks?
\newblock In Ghahramani, Z., Welling, M., Cortes, C., Lawrence, N., and
  Weinberger, K.~Q. (eds.), \emph{Advances in Neural Information Processing
  Systems}, volume~27, pp.\  3320--3328. Curran Associates, Inc., 2014.
\newblock URL
  \url{https://proceedings.neurips.cc/paper/2014/file/375c71349b295fbe2dcdca9206f20a06-Paper.pdf}.

\bibitem[Zagoruyko \& Komodakis(2016)Zagoruyko and
  Komodakis]{zagoruyko2016wide}
Zagoruyko, S. and Komodakis, N.
\newblock Wide residual networks.
\newblock In Richard C.~Wilson, E. R.~H. and Smith, W. A.~P. (eds.),
  \emph{Proceedings of the British Machine Vision Conference (BMVC)}, pp.\
  87.1--87.12. BMVA Press, 2016.

\bibitem[Zhou et~al.(2019)Zhou, Lan, Liu, and Yosinski]{zhou2019deconstructing}
Zhou, H., Lan, J., Liu, R., and Yosinski, J.
\newblock Deconstructing lottery tickets: Zeros, signs, and the supermask.
\newblock In Wallach, H., Larochelle, H., Beygelzimer, A., d\textquotesingle
  Alch\'{e}-Buc, F., Fox, E., and Garnett, R. (eds.), \emph{Advances in Neural
  Information Processing Systems}, volume~32, pp.\  3597--3607. Curran
  Associates, Inc., 2019.
\newblock URL
  \url{https://proceedings.neurips.cc/paper/2019/file/1113d7a76ffceca1bb350bfe145467c6-Paper.pdf}.

\bibitem[Zoph \& Le(2017)Zoph and Le]{zoph2016neural}
Zoph, B. and Le, Q.~V.
\newblock Neural architecture search with reinforcement learning.
\newblock In \emph{International Conference on Learning Representations}, 2017.
\newblock URL \url{https://openreview.net/pdf?id=r1Ue8Hcxg}.

\end{thebibliography}
\bibliographystyle{icml2021}

\newpage
$\:$\pagebreak
\appendix
\section{Distribution of Selected Weights and Scores}\label{additional-distributions}
\begin{figure}
    \centering
    \includegraphics[width=\linewidth]{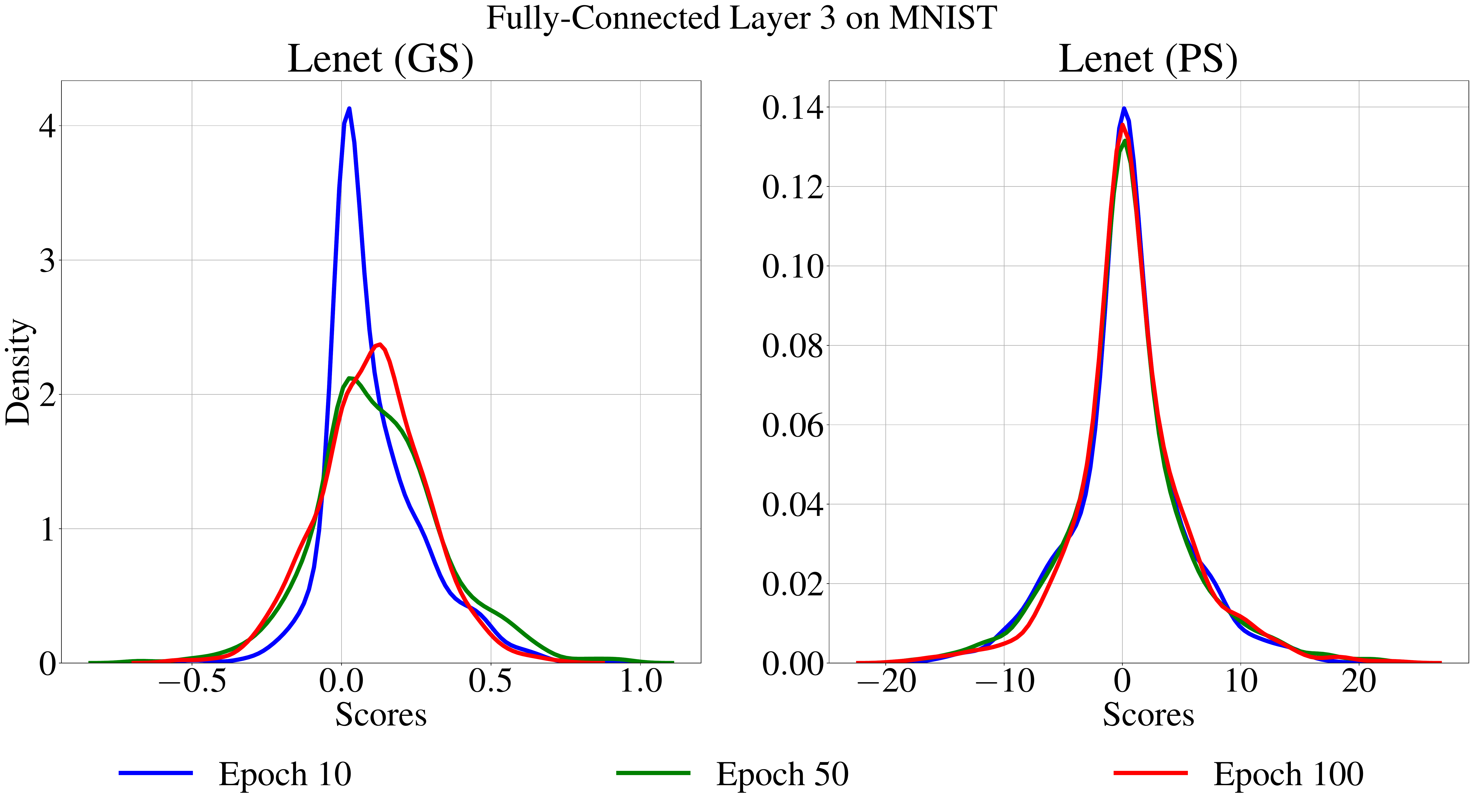}
    \caption{\textbf{Distribution of selected scores. } Different from the selected weights, the selected scores tend to be normally distributed for both $\textsc{GS}$ and $\textsc{PS}$. We show only the scores for layer $3$ of Lenet because it is the layer with the fewest number of weights. However, the other layers show a similar trend except that the selected scores in them have very narrow distributions which makes them uninteresting. Notice that although we sample the scores uniformly from the non-negative range $\mathbb{U}(0, 0.1 * \sigma_x$) where $\sigma_x$ is the standard deviation of the Glorot Normal distribution, gradient descent is able to drive them into the negative region. The scores in $\textsc{PS}$ slot machines move much farther away from the initialization compared to those in $\textsc{GS}$ due to the large learning rates used in \textsc{PS} models. }
    \label{fig:score-distrob-mnist}
\end{figure}

\begin{figure*}
    \centering
    \begin{subfigure}{\linewidth}
    \includegraphics[width=\linewidth]{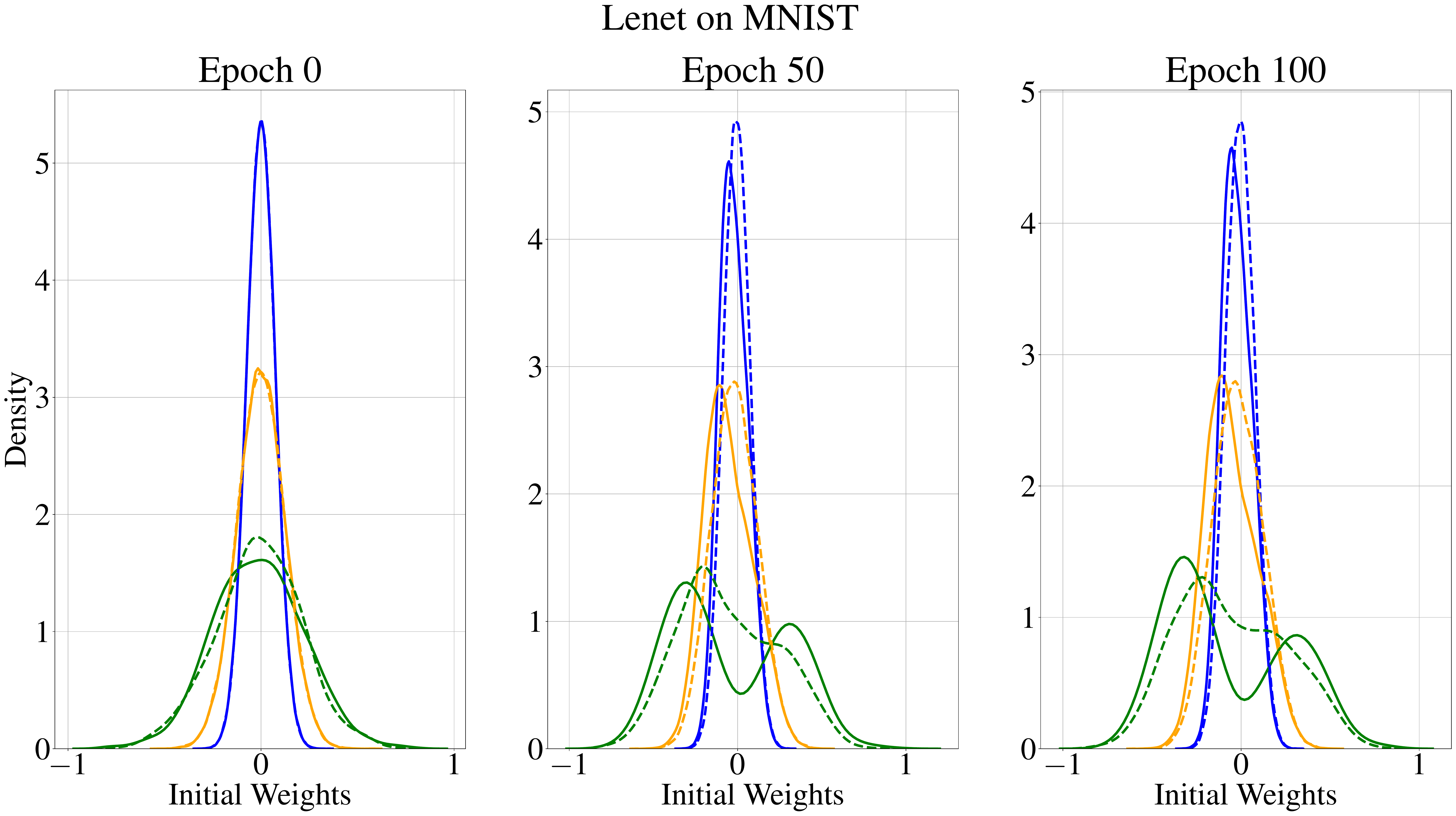}
    \caption{Glorot Normal Initialization}
    \label{glorot-normal}
    \end{subfigure}
    ~~~~
    \begin{subfigure}{\linewidth}
    \includegraphics[width=\linewidth]{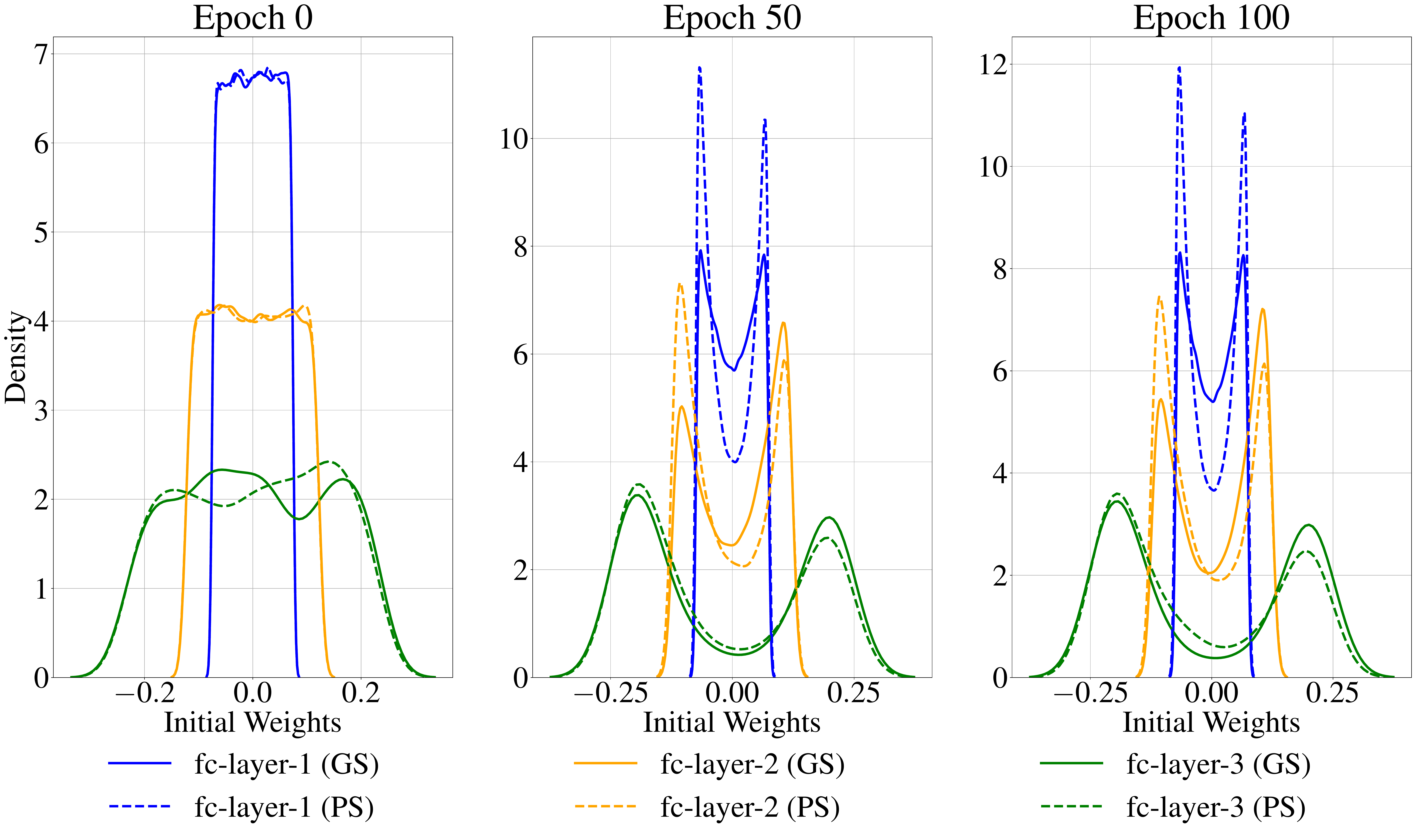}
    \caption{Glort Uniform Initialization}
    \end{subfigure}
    \caption{\textbf{Distribution of selected weights on MNIST. } As noted above, both sampling methods tend to choose larger magnitude weights as oppose to small values. This behavior is more evident when the values are sampled from a Glorot Uniform distribution (\textit{bottom}) as opposed to a Glorot Normal distribution (\textit{top}). However, layer $3$ which has the fewest number of weights of any layer in this work continue to select large magnitude weights even when using a normal distribution.}
    \label{fig:weight-distribution-mnist}
\end{figure*}

\begin{figure*}
    \centering
    \begin{subfigure}{\linewidth}
        \includegraphics[width=\linewidth]{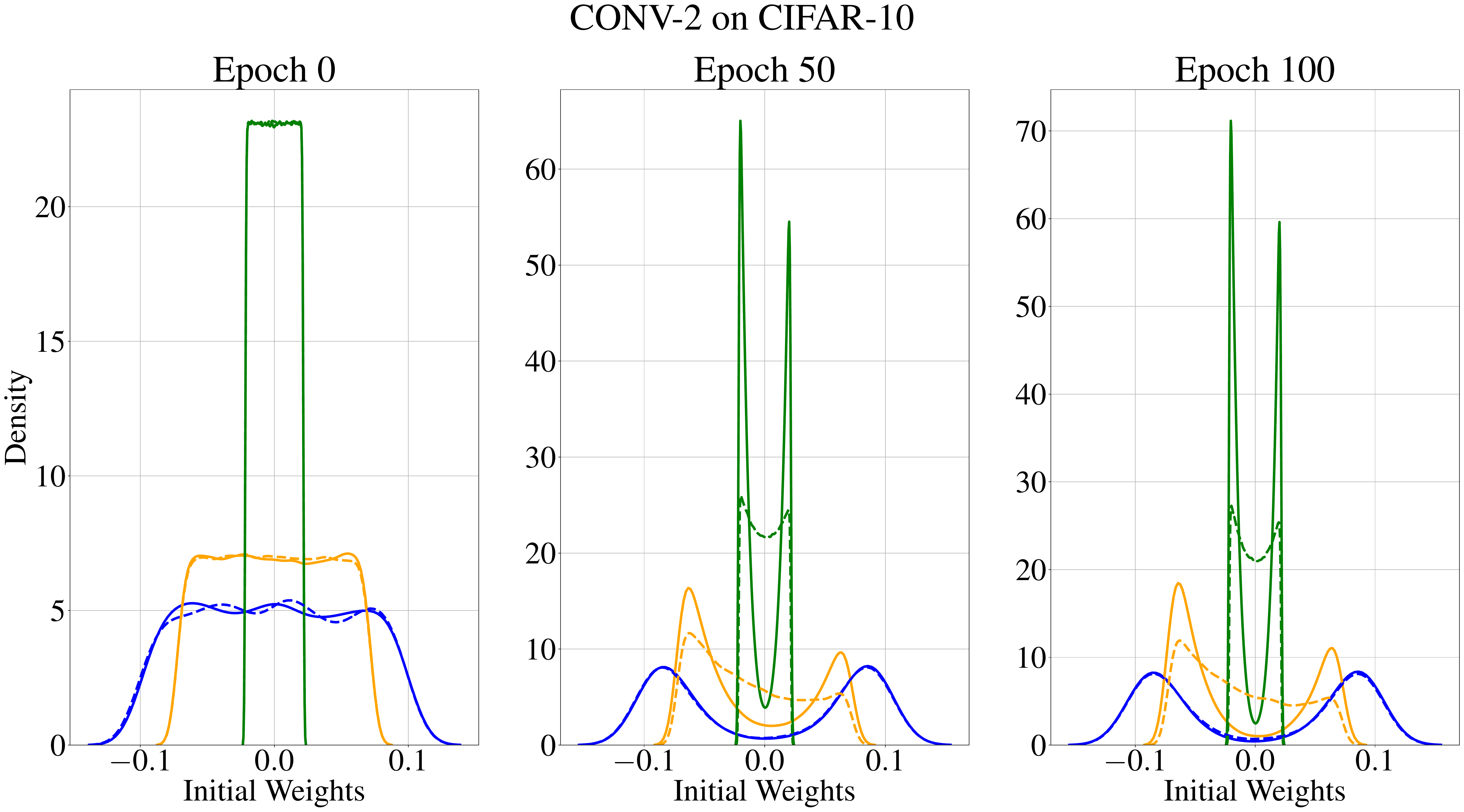}
    \end{subfigure}
    
    \vspace{0.5cm}
    
    \begin{subfigure}{\linewidth}
    \includegraphics[width=\linewidth]{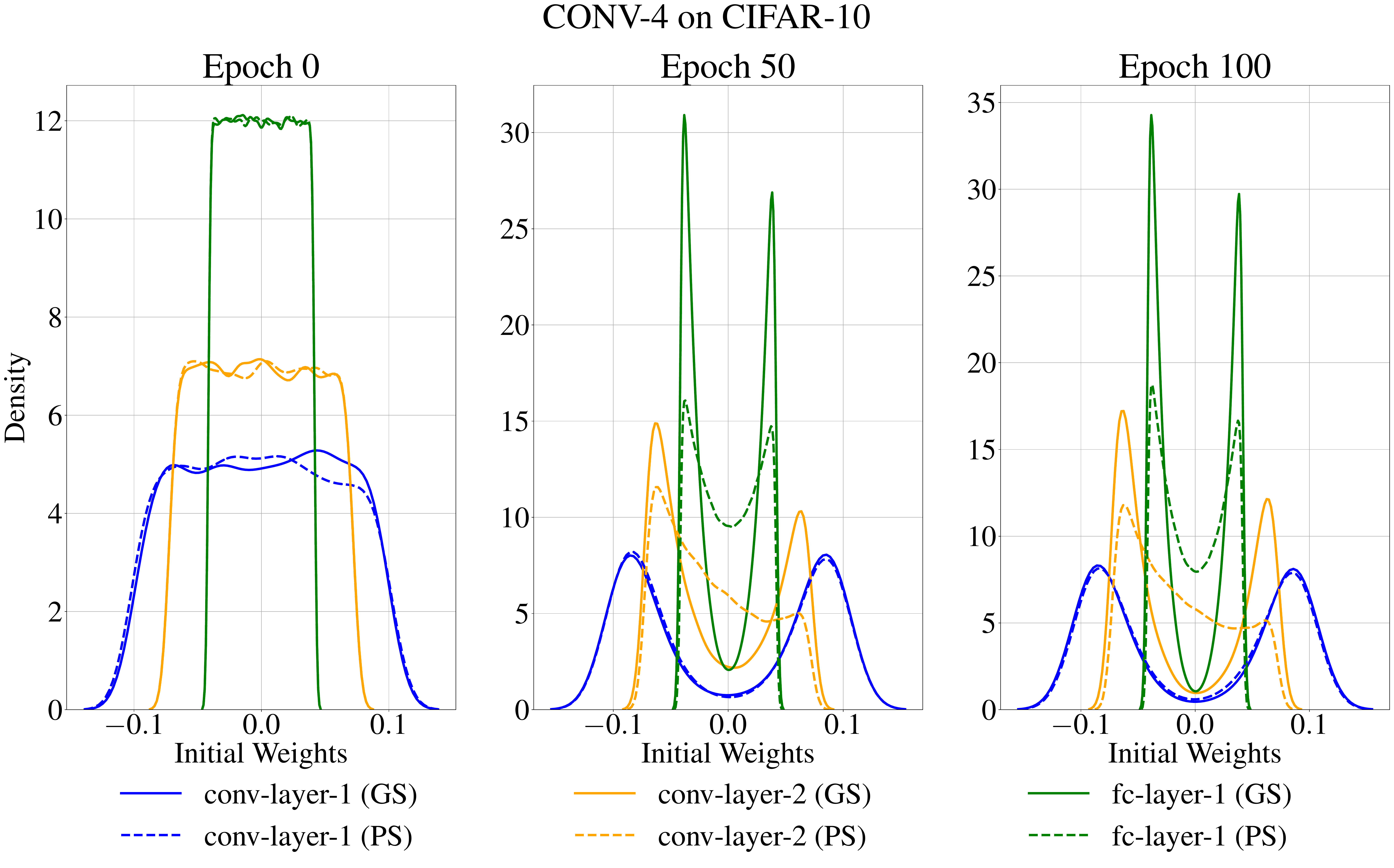}
    \end{subfigure}
    \caption{\textbf{Distribution of selected weights on CIFAR-10. } Similar to the plots shown in Figure~11 in the paper, both CONV-2 and CONV-4 on CIFAR-10 tend to choose bigger and bigger weights in terms of magnitude as training progresses. Here, we show the distribution of the selected networks in the first two convolutional layers and the first fully-connected layer of the above networks but all the layers in all slot machines show a similar pattern. }
    \label{fig:weight-distribution}
\end{figure*}

As discussed in Section~4.8 in the main paper, we observe that slot machines tend to choose increasingly large magnitude weights as learning proceeds. In Figures~\ref{fig:score-distrob-mnist},~\ref{fig:weight-distribution-mnist}, and~\ref{fig:weight-distribution} of this appendix, we provide additional plots demonstrating this phenomenon for other architectures. It may be argued that the observed behavior might be due to the Glorot Uniform distribution from which the weights are sampled. Accordingly, we performed ablations for this where we used a Glorot Normal distribution for the weights as opposed to the Glorot Uniform distribution used throughout the paper. As shown in Figure~\ref{glorot-normal}, the initialization distribution do indeed contribute to observed pattern of preference for large magnitude weights. However, initialization may not be the only reason as the models continue to choose large magnitude weights even when the weights are sampled from a Glorot Normal distribution. This is shown more clearly in the third layer of Lenet which has relatively fewer weights compared to the first two layers. We also observed a similar behavior in normally distributed convolutional layers. 

Different from the weights, notice that the selected scores are distributed normally as shown in Figure~\ref{fig:score-distrob-mnist}. The scores in $\textsc{PS}$ move much further away from the initial values compared to those in $\textsc{GS}$. This is largely due to the large learning rates used in $\textsc{PS}$ models.

\begin{figure}
    \centering
    \includegraphics[width=\linewidth]{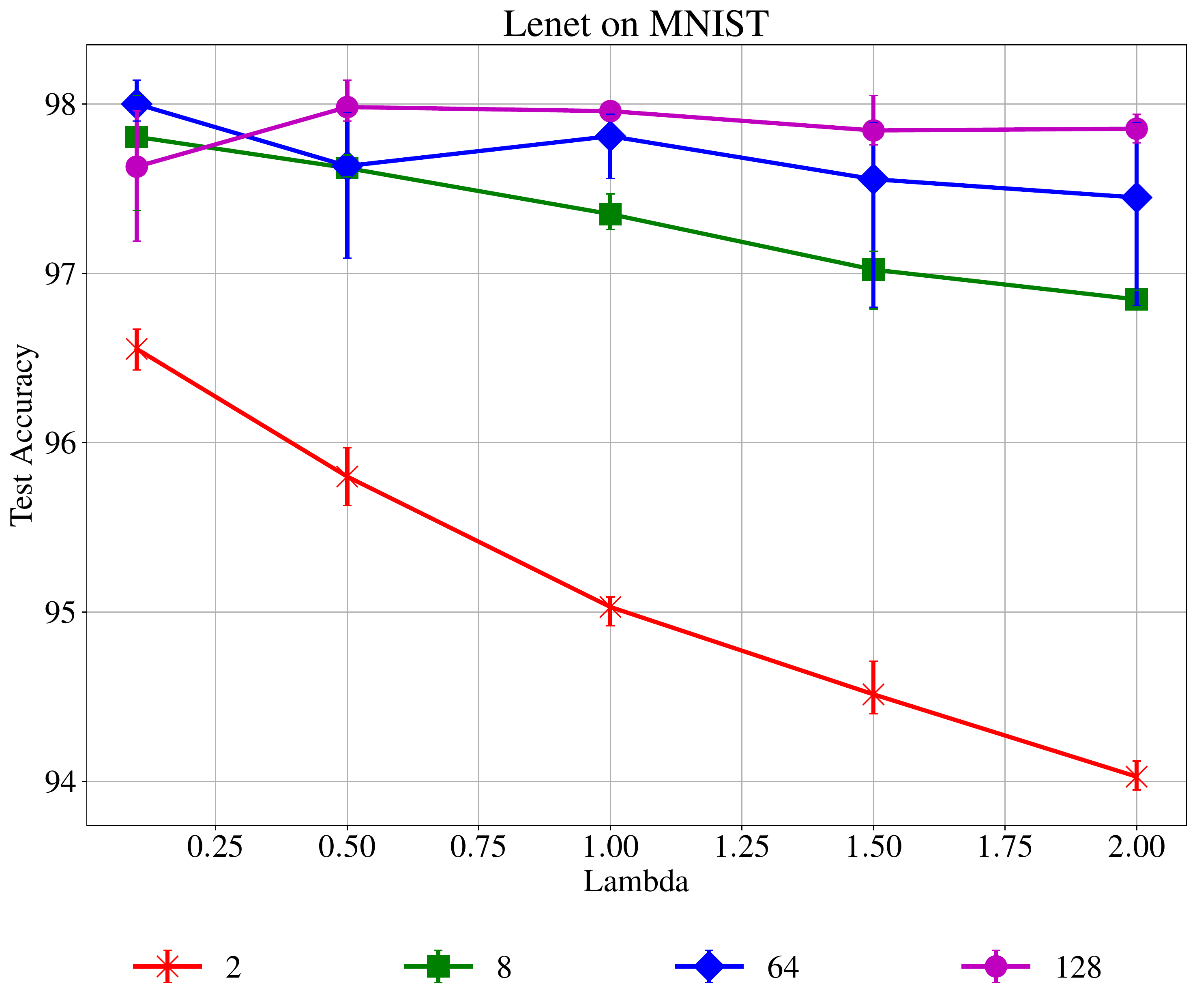}
    \caption{\textbf{Scores Initialization. } The models are sensitive to the range of the sampling distribution. As discussed in Section~4.1 of the main paper, the initial scores are sampled from the uniform distribution $\mathbb{U}(\gamma, \gamma + \lambda\sigma_x)$. The value of $\gamma$ does not affect performance and so we always set it to $0$. These plots are averages of $5$ different random initializations of Lenet on MNIST.}
    \label{fig:range-plots}
\end{figure}

\section{Scores Initialization}\label{initialization}

We initialize the quality scores by sampling from a uniform distribution $\mathbb{U}(\gamma, \gamma + \lambda\sigma_x)$. As shown in Figure~\ref{fig:range-plots}, we observe that our networks are sensitive to the range of the uniform distribution the scores are drawn from when trained using $\textsc{GS}$. However, as expected we found them to be insensitive to the position of the distribution $\gamma$. Generally, narrow uniform distributions, e.g., $\mathbb{U}(0, 0.1)$, lead to higher test set accuracy compared to wide distributions e.g., $\mathbb{U}(0, 1)$. This matches intuition since the network requires relatively little effort to drive a very small score across a small range compared to a large range. To concretize this intuition, take for example a weight $\tilde w$ that gives the minimum loss for connection ($i, j$). If its associated score $\tilde s$ is initialized poorly to a small value, and the range is small, the network will need little effort to push it to the top to be selected. However, if the range is large, the network will need much more effort to drive $\tilde s$ to the top for $\tilde w$. We believe that this sensitivity to the distribution range could be compensated by using higher learning rates for wider distributions of scores and vice-versa. 





\end{document}